\documentclass[journal]{IEEEtran}
\usepackage[square,numbers]{natbib} 
\usepackage[inline]{trackchanges}

\usepackage{multirow}
\usepackage{booktabs}
\usepackage{amsfonts} 
\usepackage{bm}
\usepackage{balance}
\usepackage{adjustbox}
\usepackage{enumitem}
\usepackage{array}
\usepackage{graphicx}
\usepackage{amsmath}
\usepackage{amsthm}
\usepackage{algorithm}
\usepackage{algcompatible}
\usepackage{url} 
\usepackage{algpseudocode}
\usepackage{amssymb}
\usepackage{caption}
\usepackage{multirow}
\usepackage{booktabs}
\theoremstyle{definition}

\usepackage{color}
\usepackage{colortbl}
\definecolor{purple}{rgb}{1,0,1}
\definecolor{black}{rgb}{0,0,0}
\definecolor{gray}{rgb}{0.75,0.75,0.75}
\usepackage{xcolor}

\definecolor{cvprblue}{rgb}{0.21,0.49,0.74}
\usepackage[pagebackref,breaklinks,colorlinks,allcolors=cvprblue]{hyperref}

\begin{document}

\title{INTENT: Trajectory Prediction Framework with Intention-Guided Contrastive Clustering}




\author{Yihong~Tang,
        Wei~Ma$^\dagger$,~\IEEEmembership{Member, IEEE}
\thanks{Yihong Tang is with McGill University, Montreal, Canada (E-mail: yihong.tang@mail.mcgill.ca). Wei Ma is with the Hong Kong Polytechnic University, Hong Kong SAR, China (E-mail: wei.w.ma@polyu.edu.hk)}
\thanks{$^\dagger$ Corresponding author.}
}

\maketitle

\begin{abstract}
Accurate trajectory prediction of road agents (e.g., pedestrians, vehicles) is an essential prerequisite for various intelligent systems applications, such as autonomous driving and robotic navigation. 
Recent research highlights the importance of environmental contexts (e.g., maps) and the "multi-modality" of trajectories, leading to increasingly complex model structures. However, real-world deployments require lightweight models that can quickly migrate and adapt to new environments. Additionally, the core motivations of road agents, referred to as their intentions, deserves further exploration.
In this study, we advocate that understanding and reasoning road agents' intention plays a key role in trajectory prediction tasks, and the main challenge is that the concept of intention is fuzzy and abstract.
To this end, we present \textsc{Intent}, an efficient intention-guided trajectory prediction model that relies solely on information contained in the road agent's trajectory.
Our model distinguishes itself from existing models in several key aspects: (i) We explicitly model road agents' intentions through contrastive clustering, accommodating the fuzziness and abstraction of human intention in their trajectories. (ii) The proposed \textsc{Intent} is based solely on multi-layer perceptrons (\textsc{Mlp}s), resulting in reduced training and inference time, making it very efficient and more suitable for real-world deployment. (iii) By leveraging estimated intentions and an innovative algorithm for transforming trajectory observations, we obtain more robust trajectory representations that lead to superior prediction accuracy.
Extensive experiments on real-world trajectory datasets for pedestrians and autonomous vehicles demonstrate the effectiveness and efficiency of \textsc{Intent}. 
\end{abstract}


\IEEEpeerreviewmaketitle



\section{Introduction}

With the rapid development of navigation systems \cite{han2017systematic,silva2018discovering,tang2023routekg} and a variety of emerging intelligent autonomous applications \cite{tang2022domain,wang2019spatiotemporal,ge2019route,yang2020context,anwar2018capturing,huang2019road,tang-etal-2024-itinera} (\textit{e.g.}, human-computer interactive systems, autonomous driving, route planning, \textit{etc.}), 
how to model the behavior of each intelligent agent is of great importance to the safe and efficient operation for these systems \cite{qu2023adversarial,mrazovic2018multi,li2019efficient,tang2024activity}.
In transport scenarios, it is essential to understand and accurately predict the short-term behaviors of road agents ({\em e.g.}, pedestrians, vehicles). Among many related research problems, the study of short-term trajectory prediction of road agents has made significant progress in recent years \cite{helbing1995social,gupta2018social,alahi2016social}. For example, \href{https://www.argoverse.org/}{Argoverse} has published two versions of motion forecasting datasets \cite{chang2019argoverse,wilson2021argoverse} and the corresponding \href{https://eval.ai/web/challenges/challenge-page/454/overview}{challenges} attracted around 260 teams with more than 2,700 submissions to solve the trajectory prediction problem and contribute to self-driving vehicle's development. As for academia, the \href{https://www.epfl.ch/labs/vita/}{VITA Lab} of EPFL held \href{https://www.aicrowd.com/challenges/trajnet-a-trajectory-forecasting-challenge}{TrajNet++} pedestrian trajectory prediction challenge to understand human behavior in large-scale scenarios. 
With accurate trajectory prediction, the safety of autonomous road agents ({\em e.g.}, autonomous cars, delivery robots) can be ensured, and hence we have good reasons to expect trajectory prediction to be one essential component in future autonomous transport systems.

\begin{figure}[t]
    \centering
    \includegraphics[width=\linewidth]{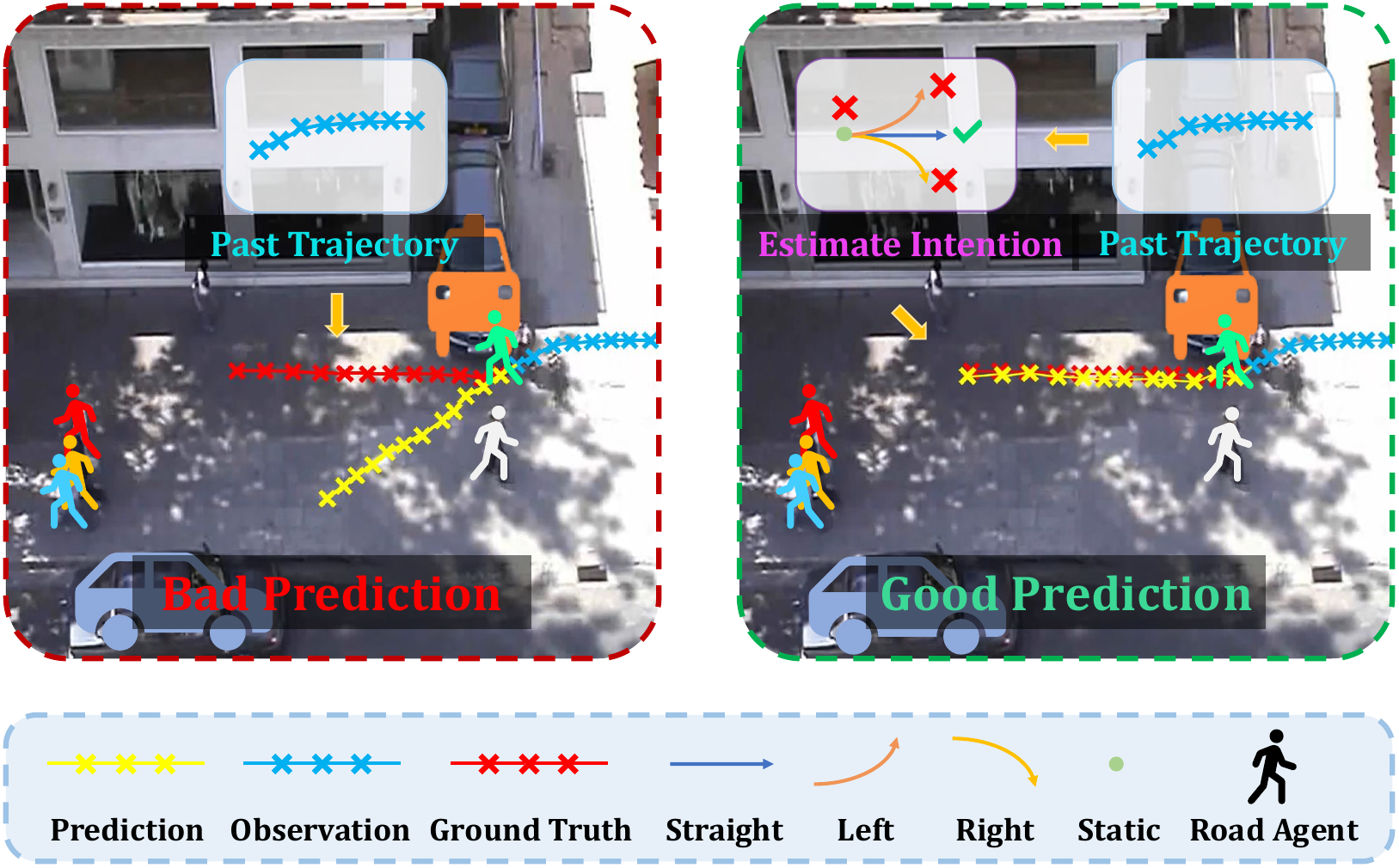}
    \caption{Illustration for intention guided trajectory prediction. The green pedestrian plans to avoid the parked orange car in order to walk forward. Existing methods may be influenced by the last few time steps of this person's past locations and make inaccurate predictions. In contrast, a trajectory prediction model guided by the estimated intention can make a more reliable and accurate prediction.}
    \label{fig:illustration}
    \vspace{-3mm}
\end{figure}

The trajectory prediction task aims to forecast road agents' future trajectories based on their observed past states. Earlier works \cite{helbing1995social,gupta2018social} produce the most likely trajectory prediction (\textit{i.e.}, deterministic prediction). \cite{helbing1995social} adopted an \textsc{Social Force} based energy function to predict a pedestrian's following action. With the rise of deep learning, numerous works  have been proposed to leverage \textsc{Rnn} based encoder and decoder to model human trajectories \cite{gupta2018social,zhang2019sr}. 
Additionally, \textsc{Social-gan} \cite{gupta2018social} introduced stochastic trajectory prediction that proposes to generate multiple predictions to model the ``multi-modality'' of human behaviors. 

Recent works enhance the understanding of the behavior of road agents by introducing some physical constraints and social interactions, such as considering context information (\textit{e.g.}, map information, spatial information, system dynamics, \textit{etc.}) \cite{lee2017desire,sadeghian2019sophie,marchetti2020mantra,tang-etal-2024-itinera} and social interaction behaviors between road agents while predicting their future trajectories \cite{mohamed2020social,kosaraju2019social}. While emerging trajectory prediction models continually emphasize social interactions, \cite{makansi2021you,saadatnejad2022sattack} questioned the contribution of social interaction mechanisms in the problem of trajectory prediction. \cite{makansi2021you}'s experimental results also demonstrate that existing SOTA models \cite{salzmann2020trajectron++,mangalam2020not,mohamed2020social} still have similar prediction performances after removing the interaction features (modules) through the feature attribution method. 
In contrast to considering context information, several methods emphasize the importance of understanding agents' goals when predicting their future trajectories \cite{mangalam2020not,zhao2020tnt}. \textsc{Pecnet} conditioned their predictions on the estimated final location of future trajectory to get more interpretable and precise predictions \cite{mangalam2020not}. However, most of these works \cite{mangalam2020not,choi2019drogon,wang2022stepwise} are stochastic and leverage estimated location goal states as precise information, which limits the practicability and applicability of these models. In real-world scenarios, the context information can change dynamically due to the deployment of various sensors. This variability, coupled with the heavy computational demands in autonomous environments, where numerous diverse scenarios must be processed, necessitates model designs that can rapidly adapt to new environments. It is crucial for models to be flexible and resilient, enabling quick adaptation to new contexts and scenarios to ensure robust performance and reliability in practical applications. This adaptability is essential for managing the extensive and varying conditions encountered in real-world deployments, ensuring models remain effective and efficient under different operational contexts.


The intention of a road agent can be defined as its subjective tendency to take a specific action or to reach a specific location, which is often seen as the internal reason for taking the corresponding action \cite{le2020contrastive}. 
Understanding the intentions of road agents is essential for trajectory prediction and robotic navigation, which can help reduce the future trajectory search space, better understand human decision-making behavior, and improve prediction accuracy. 
Figure \ref{fig:illustration} shows that a more reasonable prediction of future trajectories can be made with an accurate estimation of human intentions.
Compared to the goal-estimation methods, on one hand, accurate intention modeling provides a more meaningful and interpretable early signal, which can facilitate a variety of trajectory-based applications. On the other hand, the intention is a relatively ``fuzzy'' and abstract concept, so it is difficult to assign a specific label as in the goal-estimation methods. As a result, existing models still lack a comprehensive framework to explicitly model agents' intentions for trajectory prediction.

Some works divided drivers' intentions at intersections into different classes (i.e., left, right, straight, etc.) to enhance vehicle trajectory prediction \cite{fang2022behavioral,zyner2019naturalistic}. However, such heuristics are suitable when the trajectories of vehicles are usually regular under the constraints of lanes, and the intentions of drivers are easy to monitor and classify. Still, in scenarios with heterogeneous road agents, the trajectories of these road agents are usually complex and disorganized, and a large proportion of their trajectories cannot be simply attributed to a certain class. For example, a pedestrian's trajectory may reflect both a straight-ahead and a left-turn intention. In this case, it is difficult to assign labels to these trajectories to indicate their intention classes. Given this, contrastive learning emerges as an appropriate solution \cite{chen2020simple}, which develops metrics to guide positive pairs to be closer and push away negative pairs in the embedding space. In contrast to supervised classification or clustering methods that guide samples' representations in a top-down fashion, contrastive learning organizes samples' representations from the bottom up \cite{le2020contrastive}. Therefore we could cluster trajectories representations by constructing positive pairs across different classes to handle the ``fuzzy'' nature of road agents' intentions.

Based on the research gaps mentioned above regarding existing trajectory prediction methods and the lack of efforts in mining agents' intentions contained in past trajectories, we propose \textsc{Intent}, a novel intention-guided trajectory prediction framework that aims to jointly predict road agents' intentions and their future trajectories by solely considering information from their observed past trajectories. \textsc{Intent} emphasizes road agents' intentions when predicting their future trajectories and model the ``multi-modality'' and ``fuzzy'' nature of road agents' trajectories by four basic intention classes (i.e., straight, left, right and static, respectively) together with the cross-classes contrastive clustering technique. As a result, more intention-oriented trajectory representations and more accurate predictions can be achieved with the estimated road agents' intentions. Specifically, to obtain structured information from past trajectories, we designed a trajectory \textit{Observation Feature Extractor} to extract velocity, radian (heading orientation), and location information at each time step from past trajectories, which represent states and intentions of the road agents reflected in their past motions. In particular, when training, \textit{Observation Feature Extractor} also assigns intention labels to some regular (i.e., shape-recognizable) trajectories. Then we propose a \textit{Representation Learner} to embed these observation features into hidden representations. In addition, we also design an \textit{Intention Clusterer} to leverage the contrastive clustering technique to refine these representations by clustering different classes of trajectories' representations. After getting powerful representations of past trajectories, \textit{Location Predictor} will decode these representations together with estimated intentions to predict future trajectories. By incorporating road agents' intentions, \textsc{Intent} effectively utilizes information from past trajectories to achieve better predictions.

To the best of our knowledge, our \textsc{Intent} is the first work to model road agents' intentions in trajectory prediction problems explicitly using contrastive learning. The main contribution of \textsc{Intent} can be summarized as follows:
\begin{itemize}[leftmargin=*]
    \item We emphasize the importance of modeling the intentions of road agents to predict their trajectories and propose an intention-based trajectory prediction framework \textsc{Intent}.
    \item We design an algorithm that automatically generates intention labels for regular trajectories and leverages the contrastive clustering to handle the ``fuzzy'' nature of road agents' intentions to refine trajectories' representations.
    \item Different from existing methods to encode the location coordinates of past trajectories, \textsc{Intent} translates the trajectory information into the velocity and orientation (radian) of road agents to obtain more powerful representations. 
    \item \textsc{Intent} avoids auto-regressive model structures (such as \textsc{Rnn}s and \textsc{Transformer}s) for the trajectory prediction task and is designed solely with \textsc{Mlp}s. This lightweight design choice not only reduces the training and inference time but also retains superior performance.
    \item Our experimental results demonstrate the effectiveness and explainability of intention modeling in our \textsc{Intent} and state-of-the-art performances are achieved on pedestrian and autonomous vehicle trajectory prediction benchmarks.
\end{itemize}

The remainder of this paper is organized as follows. Section \ref{sec:relatework} reviews the related work on trajectory prediction and relevant works to road agents' intentions estimations. Section \ref{sec:pre} formulates the trajectory prediction problems, lists adopted notations and corresponding explanations for this work. Section \ref{sec:method} elaborates detailed information for our \textsc{Intent}. Experiments and results are demonstrated in section \ref{sec:exp}. We conclude the paper in Section \ref{sec:conclusion}.

\section{Related Work}
\label{sec:relatework}

We review the related literature regarding trajectory prediction and intention estimation of road agents.

\subsection{Trajectory Prediction}

Trajectory prediction tasks aim to predict road agents' future trajectories given observation parts \cite{rudenko2020human,liang2021nettraj}. Early researchers used energy-based methods or handcrafted rules to model road agents' dynamics \cite{yamaguchi2011you,antonini2006discrete,helbing1995social}. With the rise of data-driven models, researchers have begun to propose deep learning-based methods to predict trajectories of human and autonomous vehicles, accept discrete 2D spatial coordinates as input, and predict future coordinates as road agents' future locations. Some works \cite{sadeghian2019sophie,bartoli2018context,lee2017desire,su2017predicting,xue2018ss} add context features (\textit{e.g.} maps, grouping information, \textit{etc.}) to extend model input.

\begin{figure}[h]
    \centering
    \includegraphics[width=.8\linewidth]{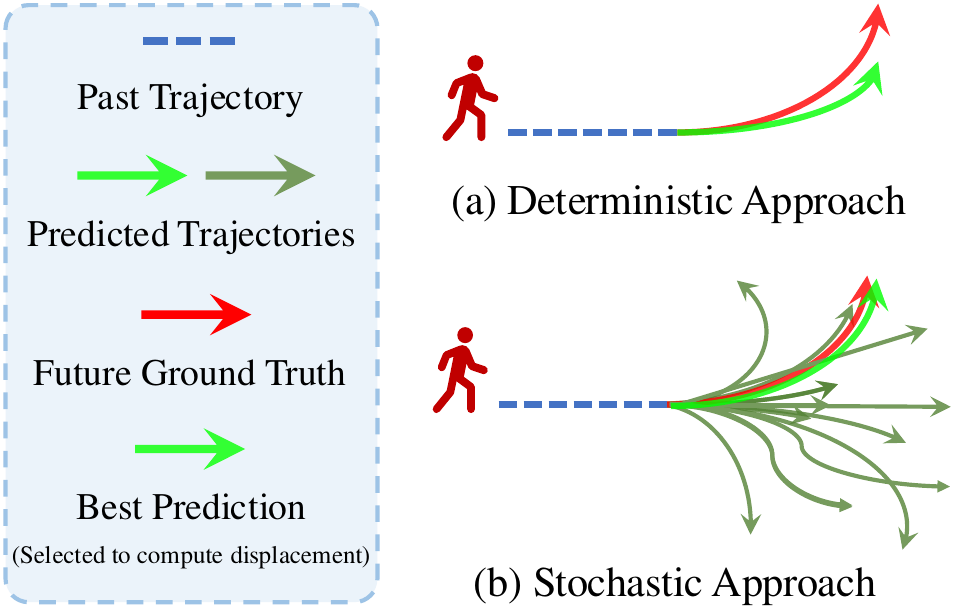}
    \caption{Differences between deterministic trajectory prediction and stochastic trajectory prediction. The deterministic method generates solely one predicted trajectory to compute displacement error while the stochastic method yields multiple predicted trajectories and selects the best prediction to compute displacement error.}
    \label{fig:ds}
\end{figure}

Existing works can be classified into two categories: deterministic methods \cite{alahi2016social} and stochastic methods \cite{gupta2018social}.
Figure \ref{fig:ds} indicates the difference between deterministic methods and stochastic methods. Predictions from stochastic methods can be of great importance for robot navigation because they produce numerous acceptable future trajectories for consideration.

Deterministic methods \cite{zhang2019sr,hochreiter1997long} consider trajectory prediction a regression task and generate only one predicted trajectory per road agent. \textsc{Social-Lstm} \cite{alahi2016social} is a pioneer work that leverage \textsc{Lstm} to encode and decode road agents' trajectories and use a social pooling module to aggregate and combine hidden states of other neighborhood road agents. \textsc{Social-Attention} \cite{vemula2018social} differs from \textsc{Social-Lstm} in that it uses social attention to gather the hidden states of neighbors. \textsc{Tf} \cite{giuliari2021transformer} uses transformer networks to predict individual road agent's future trajectory, which means \textsc{Tf} makes predictions without modeling interactions and dynamics. Still, \textsc{Tf} achieves SOTA results on the \href{http://trajnet.stanford.edu/}{TrajNet} benchmark.

Existing stochastic trajectory prediction methods \cite{gupta2018social,mangalam2020not} usually adopt Monte Carlo random sampling strategy and significantly outperform deterministic methods in displacement error metrics. These methods generate multiple (usually 20) trajectories to model the multi-modality of road agents' motions and select the predicted trajectory closest to the actual future trajectory to measure the prediction's accuracy. Generative Adversarial Network (\textsc{Gan}) based stochastic methods \cite{gupta2018social,sadeghian2019sophie,amirian2019social,kosaraju2019social,huang2020diversitygan} output various trajectories by encoding latent variables that are sampled from a simple distribution. Similar sampling methods also leverage Conditional Variation Auto Encoder (\textsc{Cvae}) to model multi-modality \cite{lee2017desire,salzmann2020trajectron++}. 



Given that the main motivation of stochastic methods is to generate multiple trajectories to model the ``multi-modality'' nature 
of the future trajectories of road agents, the phenomenon that the prediction accuracy of the stochastic methods far exceeds that of the deterministic methods call into question the appropriateness of using displacement error (\textit{e.g.} \textit{ADE}, \textit{FDE}) to compare between the stochastic and deterministic methods.
\cite{giuliari2021transformer} propose an individual-based deterministic  trajectory prediction \textsc{Transformer} model \textsc{Tf} and yields the best score on the trajectory prediction benchmark TrajNet\footnote{\url{http://trajnet.stanford.edu/}}. This demonstrates that deterministic methods can still achieve superior performance when measured in real-world scenarios, and that the prediction results of deterministic methods are more stable than those of stochastic methods. The experimental results of \textsc{Tf} illustrate the importance of deterministic methods design for trajectory prediction tasks.

\subsection{Intention Estimation}

The concept of intention is extensively utilized in the computer vision and robotics literature for action classification or path refinement \cite{rasouli2019pie}. Bandyopadhyay et al. view road agents' intentions as potential goals to refine predictions \cite{bandyopadhyay2013intention}. Several methods condition each predicted trajectory on goals to achieve more explainable and accurate predictions \cite{mangalam2020not,zeng2021lanercnn,zhao2020tnt}. However, these approaches predominantly consider location information as the goal state of the predicted trajectory. Schneemann and Heinemann detect pedestrian crossing intentions to enhance situational awareness in autonomous environments \cite{schneemann2016context}. Moreover, datasets with intention labels, such as JAAD \cite{rasouli2017they}, STIP \cite{liu2020spatiotemporal}, and PIE \cite{rasouli2019pie}, have been proposed to study pedestrians' intentions. Nevertheless, these datasets primarily focus on pedestrian intentions at intersections. To address these limitations, the LOKI dataset provides detailed, frame-wise annotations for both long-term positional goals and short-term intended actions \cite{girase2021loki}.
While the aforementioned datasets facilitate the construction of labels to measure road agents' intentions, they necessitate a substantial deployment of sensors. Furthermore, existing research has not fully explored the possibility of learning and understanding road agents' intentions solely from their trajectories due to the diverse and complex nature of these trajectories \cite{bandyopadhyay2013intention}.

\subsection{Contrastive Clustering}

Contrastive learning has recently become a dominant component in self-supervised learning tasks \cite{liu2021self} due to its success in computer vision \cite{chen2020simple}, natural language processing \cite{kim2021self,fang2020cert}, and graph learning \cite{liu2022graph} domains. It aims at learning close representations for positive pairs in the representation space while trying to push away representations for negative pairs. Recent research has considered contrastive learning in clustering \cite{li2020prototypical,caron2020unsupervised,huang2021deep,li2021contrastive,tsai2020mice}, and these works usually bring together a clustering objective with contrastive learning (i.e., conducting both instance-level and cluster-level contrastive learning) to learn more stable and robust representations. Different from existing methods, we use contrastive learning to cluster representations of trajectories across different intention classes, which is more suitable for modeling road agents' complex and ``fuzzy'' intentions.

\section{Preliminaries}
\label{sec:pre}

\renewcommand{\arraystretch}{1}
\begin{table}[h]
\caption{The Adopted Symbols and Descriptions.} 
\centering 
\normalsize
\resizebox{\linewidth}{!}{
\begin{tabular}{c|c}
\toprule Symbol & \multicolumn{1}{c}{ Description } \\
\midrule

$T$ & The complete trajectory of a road agent \\ 
$X$ & The observation trajectory of a road agent \\  
$Y$; $\widehat{Y}$ & The future trajectory of a road agent\\   
$\widehat{Y}$ & The predicted future trajectory of a road agent \\ 
$T^\prime$; $X^\prime$; $Y^\prime$; $\widehat{Y}^\prime$ & Transformed trajectories \\ 
$L_t$ & The location $(x_t, y_t)$ of a road agent at time $t$ \\ 
$L_t^\prime$ & The transformed location at time step $t$ \\ 
$n$ & Time steps of the complete trajectory \\ 
$t_{obs}$ & Time steps of the observation trajectory \\  
$t_{pred}$ &  Time steps of the future trajectory \\ 
$\gamma$ & The complete study time period \\ 
$\gamma_{obs}$ & The observation time period \\ 
$\gamma_{pred}$ & The future time period \\ 
$N$ & Num of intention classes \\ 
$D^v$ & The processed velocity sequence \\  
$D^r$ & The processed radian sequence \\ 
$D^y$ & The transformed $y$ sequence \\ 
$I$ & The intention label of a road agent \\ 
$\widehat{I}$ & The predicted intention label of a road agent \\ 
$\widehat{I_S}$ & The predicted soft intention label of a road agent \\ 
$F$ & Encoded observation feature \\ 
$R_h$ & Learned representation \\ 
$R_z$ & Projection of $R_h$ \\ 
$\textbf{MLP}$;$\textbf{Rep}$;$\textbf{Proj}$ & Model parameters \\ 
$\mathcal{L}_c$;$\mathcal{L}_r$;$\mathcal{L}_p$ & Loss terms \\ 
$\mathcal{T}$ & Hyper-parameter temperature \\ 
$\alpha$; $\beta$ & hyper-parameters trade-off different loss terms \\ 

\bottomrule
\end{tabular}
}
\label{table:notations}
\end{table}

In this section, we first present definitions relevant to our work, then formulate the problem settings of trajectory prediction. Notations used in the formulation are summarized in Table~\ref{table:notations}.

\textbf{Location.} 
The location $L_t$ of a road agent at time step $t$ is defined as a 2D coordinate $L_t=(x_t, y_t)$ and we have $L_{t}^{x}=x_t$, $L_{t}^{y}=y_t$.

\textbf{Trajectory.}
The trajectory $T$ of a road agent is defined as $n$ consecutive time steps' locations: $\{L_1, L_2, \dots, L_n\}$, where $\{\cdot\}$ is a set whose elements are ordered by time. In trajectory prediction task settings, a trajectory can be divided into an observation trajectory $X$ with $t_{obs}$ locations and a future trajectory $Y$ with $t_{pred}$ locations. Formally, $X=\{L_1, \dots, L_{t_{obs}}\}$, $Y=\{L_{t_{obs}+1}, \dots, L_{t_{obs}+t_{pred}}\}$, where 
$t_{obs}+t_{pred} = n$. For simplicity, we name $T$, $X$ and $Y$ as the complete trajectory, the observation trajectory and the future trajectory, respectively.

\textbf{Trajectory Prediction.}
Given the observation trajectory $X$ for a road agent, the trajectory prediction task aims to learn a mapping $f$ that maps $X$ to the predicted future trajectory $\widehat{Y}$, which can be specified as a mapping: $\widehat{Y}=\mathcal{M}_f(X, \theta_f)$, where $\theta_f$ is the parameters which represents the collection of all trainable weights of the neural network for the mapping $\mathcal{M}_f$.

In the following sections, we consider the study time period $\gamma=\{1, 2, \dots, t_{obs}, \dots, t_{obs}+t_{pred}\}$ for illustration. We use notation $\gamma_{obs}$ and $\gamma_{pred}$ to denote the observation time period $\{1, 2, \dots, t_{obs}\}$ and the future time period $\{t_{obs}+1, \dots, t_{obs}+t_{pred}\}$, respectively. $t_i$ represents the $i$th time step in a trajectory.

\section{Methodology}
\label{sec:method}

\begin{figure}[h]
    \centering
    \includegraphics[width=\linewidth]{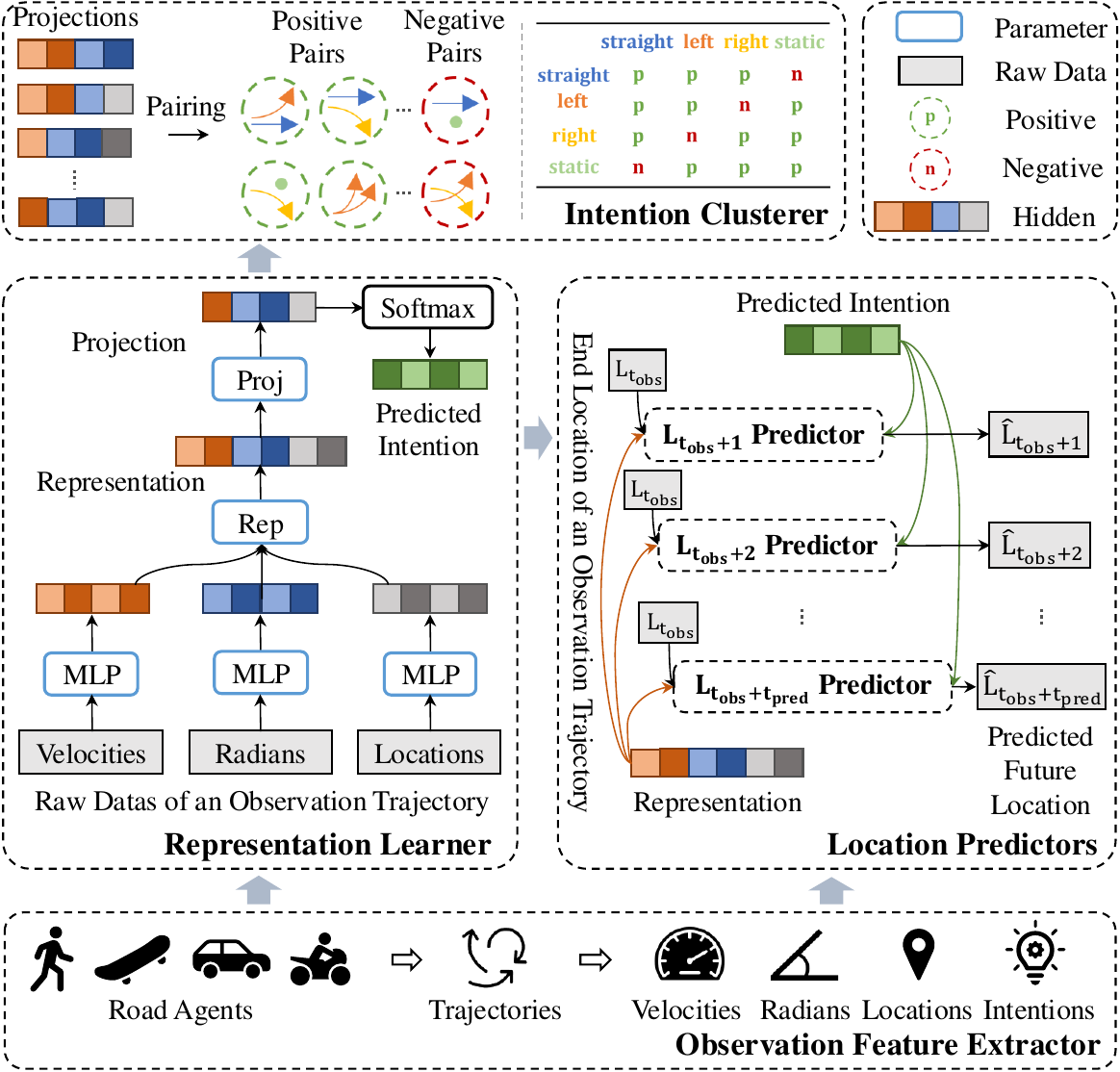}
    \caption{An overview of $\textsc{Intent}$ architecture.}
    \label{fig:model}
\end{figure}


The key idea of the proposed model is that the movements of road agents are guided by their intentions. We want to attribute and reason features from the observation trajectory that impact road agents' decisions and path plannings to predict their future trajectories better. However, the intention is a rather abstract concept with inherent ``multi-modality'' therefore challenging to model and quantify. 
To address this, we classify the intentions of road agents into $N=4$ basic intention classes: straight, left, right, and static, used to cluster the intentions of road agents.

To achieve the above goals and make accurate predictions, we design four key modules for $\textsc{Intent}$, which include an \textit{Observation Feature Extractor}, a \textit{Representation Learner}, an \textit{Intention Clusterer}, and a \textit{Location Predictor}. 

The \textit{Observation Feature Extractor} is non-parametric and it pre-processes a complete trajectory $T$ to generate the end location $L_{t_{obs}}$ of the corresponding observation trajectory $X$ and observation features. Observation features consist of velocity sequence ${D}^v$, radian sequence ${D}^r$, transformed $y$ sequence ${D}^y$ and an intention label ${I}$ in one-hot format for the observation trajectory. We use a mapping $\mathcal{M}_P$ to denote this feature extracting procedure, \textit{i.e.} $(L_{t_{obs}}, {D}^v, {D}^r, {D}^y, {I})=\mathcal{M}_P(T)$. $\mathcal{M}_P$ will be converted to the form  $(L_{t_{obs}}, {D}^v, {D}^r, {D}^y)=\mathcal{M}_P(X)$ during the evaluation stage, and it's detailed operations are explained in section \ref{subsec:preprocessor}.

Making use of the observation features $({D}^v, {D}^r, {D}^y)$, the \textit{Representation Learner} aims to learn a powerful representation $R_h$ that possesses the information of the observation trajectory $X$. The function of the \textit{Representation Learner} can be summarized as a mapping: $(R_h, \widehat{I_S}, \widehat{I}) = \mathcal{M}_R(D^v, D^r, D^y; \theta_R)$, where $\theta_R$ is parameter of mapping $\mathcal{M}_R$, $\widehat{I}$ is the predicted intention label and $\widehat{I_S}$ is the soft intention label of the observation trajectory. Detailed definitions of the two labels will be discussed later. 
The \textit{Representation Learner} minimizes the intention labels' classification loss $\mathcal{L}_r$, and hence trajectories in the same intention class could gain similar representations.

In addition, $\textsc{Intent}$ emphasizes the clustering characteristics of road agents' trajectories representations across intention classes. This leads to our \textit{Intention Clusterer}, which yields the cluster loss $\mathcal{L}_c$. Our \textit{Intention Clusterer} is also non-parametric and by descending $\mathcal{L}_c$, the parameter $\theta_R$ in mapping $\mathcal{M}_R$ is updated adversarially.

Finally, given the observation trajectory's end location $L_{t_{obs}}$ and the learned representation $R_h$, \textit{Location Predictor}s are responsible for mapping $L_{t_{obs}}$ and $R_h$ to $L_{t_{obs}+1}, \dots, L_{t_{obs}+t_{pred}}$, which can be denoted by $t_{pred}$ mappings. The mapping  predicts the future location $\widehat{L_{t}}$ of the road agent at time step $t$, as illustrated by: $\widehat{L_{t}}=\mathcal{M}_{L_{t}}(L_{t_{obs}}, R_h, \widehat{I_S}; \theta_{L_{t}})$, where $t \in \{t_{obs}+1, \dots, t_{obs}+t_{pred}\}$ and $\theta_{L_{t}}$ is corresponding parameters of $\mathcal{M}_{L_{t}}$. The predicted future trajectory $\widehat{Y}$ of the road agent is composed by ouptuts of mappings $\mathcal{M}_{L_{t_{obs}+1}}, \dots, \mathcal{M}_{L_{t_{obs}+t_{pred}}}$. We use loss $\mathcal{L}_t$ to denote the differences between predicted and actual future trajectories.

In the following sections, we elaborate on the details of the four components.

\subsection{Observation Feature Extractor}
\label{subsec:preprocessor}

Given the past trajectory $X$ of a road agent, the goal of the \textit{Observation Feature Extractor} is to generate features like velocity sequence $D^v$, radian sequence $D^r$, transformed $y$ sequence $D^y$ and the intention label $I$.

Velocity sequence $D^v$ and radian sequence $D^r$ are defined over $X$'s elements (locations). Mathematically, we denote the velocity of the road agent at time step $t \in \gamma_{obs}$ as:

\begin{equation}\label{eq1}
D^{v}_t=
    \begin{cases}
        \frac{1}{\Delta_t}\left\|L_{t+1}-L_{t}\right\|_{2}, & t \neq t_{obs}\\
        D^{v}_{t-1}, &t = t_{obs}
    \end{cases},
\end{equation}

\begin{equation}\label{eq2}
    D^{v} = \{D^{v}_t \text{ }|\text{ } t \in \gamma_{obs} \},
\end{equation}
where $\Delta_t$ indicates the sampling time interval between two consecutive locations. 
Similarly, the radian sequence $D^r$ of the road agent at time step $t \in \gamma_{obs}$ is computed by the following equations:


\begin{equation}\label{eq3}
    rad_{t}=\text{arcsin}\left(\frac{L_{t+1}^{y}-L_{t}^{y}}{\left\|L_{t+1}-L_{t}\right\|_{2}}\right),
\end{equation}
\begin{equation}\label{eq4}
    D^{r}_{t}=
        \begin{cases}
            rad_{t}, & L_{t+1}^{x} > L_{t}^{x} \wedge t \neq t_{obs} \\
            \pi-rad_{t}, & L_{t+1}^{x} \leqslant L_{t}^{x} \wedge L_{t+1}^{y} > L_{t}^{y} \wedge t \neq t_{obs} \\
           -rad_{t}-\pi, & L_{t+1}^{x} \leqslant L_{t}^{x} \wedge L_{t+1}^{y} \leqslant L_{t}^{y} \wedge t \neq t_{obs} \\
           D^r_{t_{obs}-1}, & t = t_{obs} \\
        \end{cases},
\end{equation}
\begin{equation}\label{eq5}
    D^r=\{D^{r}_{t} \text{ }|\text{ } t \in \gamma_{obs} \}.
\end{equation}


To identify the intentions of road agents from their observation trajectories, we need to transform  trajectories that are irregularly distributed within the 2D coordinate system into a more consistent distribution. To be more specific, given a complete trajectory $T$ of a road agent and time $t \in \gamma$, we first calculate the radian $\varphi$ to rotate the complete trajectory $T$:
\begin{figure}[t]
    \centering
    \includegraphics[width=\linewidth]{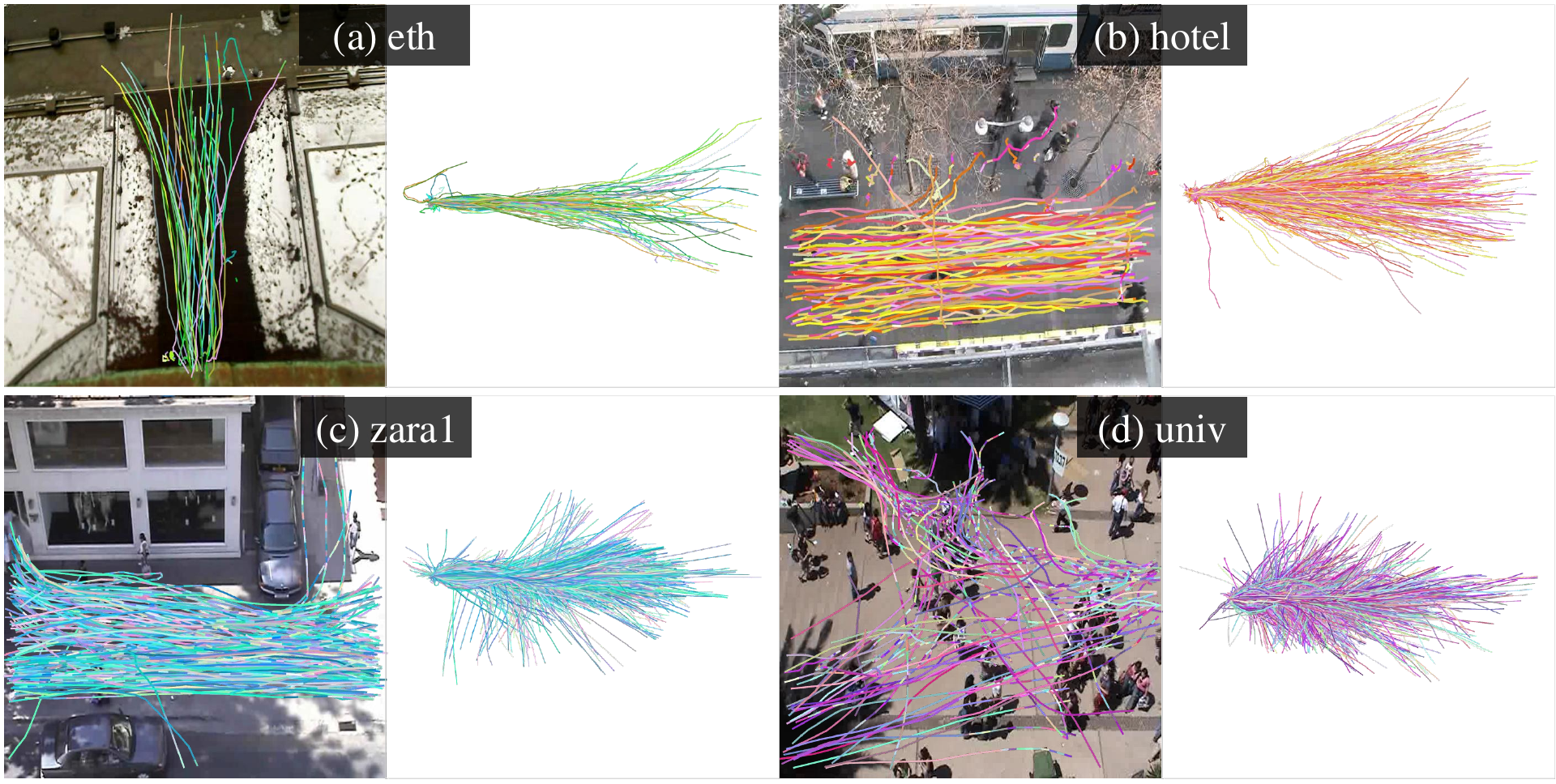}
    \caption{Visualization of transformed trajectories of the ETH-UCY dataset.}
    \label{fig:rotate}
\end{figure}


\begin{equation}\label{eq6}
    \begin{aligned}
        \varphi&=\text{ }\text{abs}\left(\text{arctan}\left(\frac{L_{t_{obs}}^{y}-L_{1}^{y}}{L_{t_{obs}}^{x}-L_{1}^{x}}\right)\right),\\
        \varphi&=
        \begin{cases}
            \frac{\pi}{2}, & \text{if} L_{t_{obs}}^{x}-L_{1}^{x} = 0 \\
            \varphi, & \text{if } L_{t_{obs}}^{x}-L_{1}^{x} > 0\\
            \pi-\varphi, & \text{otherwise }  \\
        \end{cases}.
    \end{aligned}
\end{equation}

After obtaining the rotation radian $\varphi$, we process the complete trajectory $T$ to get the transformed complete trajectory $T^\prime$ as follows:

\begin{equation}\label{eq7}
    \begin{aligned}
        x^{c}_{t}, \text{ } x^{s}_{t}\text{ }&=\text{ }\text{cos }(\varphi)\cdot(L^{x}_{t}-L^{x}_{1}),\text{ sin }(\varphi)\cdot(L^{y}_{t}-L^{y}_{1}), \\ 
        y^{c}_{t}, \text{ } y^{s}_{t}\text{ }&=\text{ }\text{sin }(\varphi)\cdot(L^{x}_{t}-L^{x}_{1}), \text{ cos }(\varphi)\cdot(L^{y}_{t}-L^{y}_{1}),
    \end{aligned}
\end{equation}

\begin{equation}\label{eq8}
     (L_{t}^{x^{\prime}}, L_{t}^{y^{\prime}})=
        \begin{cases}
            (x^{c}_{t}+x^{s}_{t}, -y^{c}_{t}+y^{s}_{t}), & \text{if} L_{t_{obs}}^{y}-L_{1}^{y}>0\\
            (x^{c}_{t}-x^{s}_{t}, y^{c}_{t}+y^{s}_{t}), & \text{otherwise}
        \end{cases},
\end{equation}

\begin{equation}\label{eq9}
    \begin{aligned}
        L^{\prime}_{t}&=(L_{t}^{x^{\prime}}, L_{t}^{y^{\prime}}) ,\\
        L^{x^\prime}, L^{y^\prime}&=\text{ }\{L_{t}^{x^{\prime}} |\text{ } t \in \gamma\}, \text{ }\{L_{t}^{y^{\prime}} |\text{ } t \in \gamma\},
    \end{aligned}
\end{equation}

\begin{equation}\label{eq10}
    X^\prime, Y^\prime, T^\prime=\text{ } \{L^{\prime}_t |\text{ } t \in \gamma_{obs}\}, \{L^{\prime}_t |\text{ } t \in \gamma_{pred}\}, \{L^{\prime}_t |\text{ } t \in \gamma\},
\end{equation}
where $X^\prime$, $Y^\prime$, $T^\prime$ represent the transformed observation trajectory, the transformed future trajectory, and the transformed complete trajectory of the road agent. Figure \ref{fig:rotate} shows a visualization of trajectories transformation results on the ETH-UCY dataset.

After obtaining the transformed observation trajectory $X^\prime=\{(L_{t}^{x^{\prime}}, L_{t}^{y^{\prime}})|t \in \gamma_{obs}\}$ of the road agent, the transformed $y$ sequence $D^{y}$ can be calculated directly by formula:
\begin{equation}\label{eq11}
    D^{y}=\{L^{y^\prime}_{t}\text{ }|\text{ } t \in \gamma_{obs}\}.
\end{equation}

Now, we are ready to assign intention labels to the trajectories. Road agents exhibit a wide range of behaviors, making their trajectories diverse and often unpredictable. Due to this irregularity, we can only assign intention labels to smoother trajectories. To start, we define a function that indicates the variation trend of a sequence. Note that we use the complete trajectory for computation, as we only assign labels to trajectories in the training set:
\begin{equation}\label{eq12}
    \mathcal{I}^+(x)=\sum_{i=2}^{t_{obs}+t_{pred}}1_{\mathbb{R}_{<0}}(x_i-x_{i-1}),
\end{equation}
where $1_{\mathbb{R}_{<0}}(\cdot) \in \{0, 1\}$ 
is an indicator function:
\begin{equation}\label{eq13}
    1_{\mathbb{R}_{<0}}(x) = 
    \begin{cases}
        0, & \text{if } x \geqslant 0 \\
        1, & \text{otherwise} \\
    \end{cases}.
\end{equation}
$\mathcal{I}^+(x)$ calculates the number of times the value of the $x$ decreases in the sequence, and we make the function reflect the increasing trend of $x$ sequence by restricting its value to smaller than a specific threshold.

Then the ground true intention label to a trajectory is computed based on the formula:
\begin{equation}\label{eq14}
    I=\textbf{GetIntentionLabel}(L^{x^\prime}, L^{y^\prime}, D^v, D^r).
\end{equation}

The function $\textbf{GetIntentionLabel}$ is defined in Algorithm \ref{alg}, where $\text{mean}(\cdot)$ and $\text{std}(\cdot)$ calculate the mean value and standard deviation of a given input sequence. 
We denote $\text{thresh}_y$, $\text{thresh}_v$, $\text{V}_s$, $\text{V}_l$, $\text{V}_r$, and $\text{V}_a$ as hyper-parameters that may vary due to changes in the dataset.  In general, $\textbf{GetIntentionLabel}$ determines the intention of a road agent by identifying variation trends of its transformed trajectory coordinates $L^{x^\prime}, L^{y^\prime}$, the velocity sequence $D^v$, and the radian sequence $D^r$, and assigns the intention label accordingly.

\begin{algorithm}[h]
\caption{The algorithm of assigning intention label.}\label{alg}
\begin{algorithmic}
\State \hspace{-0.125in}\textbf{GetIntentionLabel}(Transformed $x$ sequence $L^{x^\prime}$, Transformed $y$ sequence $L^{y^\prime}$, Velocity sequence $D^v$, Radian sequence $D^r$)

\State set hyper-parameters $\text{thresh}_v$, $\text{thresh}_y$, $\text{V}_a$, $\text{V}_{s}$ and $\text{V}_{lr}$
\If{mean($D^v$)  $< \text{V}_a$}
\State {\bf return} static
\ElsIf{mean($D^v$) $ > \text{V}_s$ \textbf{and} std($L^{y^\prime}$) $< \text{thresh}_y$ \textbf{and} \\ \hspace{0.38in} $\mathcal{I}^+(L^{x^\prime})  \leqslant \text{thresh}_v$}
\State {\bf return} straight
\ElsIf{mean($D^v$) $ > \text{V}_{lr}$ \textbf{and} std($L^{y^\prime}$) $< \text{thresh}_y$ \textbf{and} \\ \hspace{0.38in} $\mathcal{I}^+(L^{x^\prime}) \leqslant \text{thresh}_v$}
    \If{$\mathcal{I}^+(-L^{y^\prime}) \leqslant \text{thresh}_v$ \textbf{and} $\mathcal{I}^+(-D^{r}) \leqslant \text{thresh}_v$}
    \State {\bf return} right
    \ElsIf{$\mathcal{I}^+(L^{y^\prime}) \leqslant \text{thresh}_v$ \textbf{and} $\mathcal{I}^+(D^{r}) \leqslant \text{thresh}_v$} 
    \State {\bf return} left
    \EndIf
\EndIf
\end{algorithmic}
\end{algorithm}

Finally, the complete trajectory $T$ is sliced to obtain the final observation location $L_{t_{obs}}$. Mapping $\mathcal{M}_p$ is fulfilled by above described procedures. Note that during evaluation stage of $\textsc{Intent}$, we only input the observation trajectory $X$ to generate 
$D^v$, $D^r$, $D^y$, $L_{t_{obs}}$, and the intention labels are not provided.

\subsection{Representation Learner}
\label{subsec:Repextract}

Motion patterns of road agents are reflected in their trajectories. Our proposed \textit{Representation Learner} learns the representation that helps predict road agents' intentions, extracts features related to its motion planning, and removes redundant information. Meanwhile, we use intention labels to guide \textit{Representation Learner} to generate similar representations for trajectories that belong to the same intention class.

As defined in mapping $\mathcal{M}_R$, given a road agent's observation features $D^v$, $D^r$ and $D^y$, the \textit{Representation Learner} maps these features to the road agent's observation trajectory representation $R_h$ and predict the road agent's intention label $\widehat{I}$ as well as it's soft intention label $\widehat{I_S}$ which demonstrate the probabilities that the road agent's intention belong to each intention class. To realize this, we first encode processed data to unified feature $F$ with data-specific multilayer perceptron (\textbf{MLP}) as:

\begin{equation}\label{eq15}
F = \text{concat}\left(\textbf{MLP}^{s}(D^v), \textbf{MLP}^{r}(D^r), \textbf{MLP}^{y}(D^y) \right)
\end{equation}

Inspired by the contrastive learning framework \cite{chen2020simple}, we use a base encoder $\textbf{Rep}$ that extracts representation vector $R_h$ from the observation feature: 
\begin{equation}\label{eq16}
    R_h = \textbf{Rep}(F),
\end{equation}
then a projection encoder $\textbf{Proj}$ that maps $R_h$ to the vector space where cluster loss $\mathcal{L}_c$ applied: 
\begin{equation}\label{eq17}
    R_z = \textbf{Proj}(R_h).
\end{equation}
We use $R_z$ to denote the output projection from $\textbf{Proj}$, $R_z$'s dimension equals to $N$. Finally, $R_z$ is mapped to the soft label $\widehat{I_S}$ through a $\text{softmax}$ layer. This process is implemented as:
\begin{gather}\label{eq18}
    \begin{aligned}
        \widehat{I_S} &= \text{softmax}(R_z),\\
        \widehat{I}&= \arg\max(\widehat{I_S}).
    \end{aligned}
\end{gather}
For simplicity, both $\textbf{Rep}$ and $\textbf{Proj}$ are implemented as $\textbf{MLP}$.

The classification loss $\mathcal{L}_r$ function is defined as a cross-entropy term as follows:
\begin{equation}\label{eq19}
    \mathcal{L}_r=-\sum^{N}_{c=1}I[c]\log\left(\widehat{I_S}[c]\right),
\end{equation}
where $\widehat{I_S}[c]$ denotes the $c$th value in $\widehat{I_S}$ and $I[c]$ denotes the $c$th true intention label.

\subsection{Intention Clusterer}
\label{subsec:cluster}

The intentions of road agents are complex and diverse, so a simple classification and labeling of trajectories may not accurately model their intentions. Therefore, we expect ``similar'' trajectories to have clustered representations, even though they may not belong to the same intention class. To achieve this, we introduce our \textit{Intention Clusterer}, which aims to contrastively cluster the representations for similar trajectories.

\begin{figure}[h]
    \centering
    \includegraphics[width=.85\linewidth]{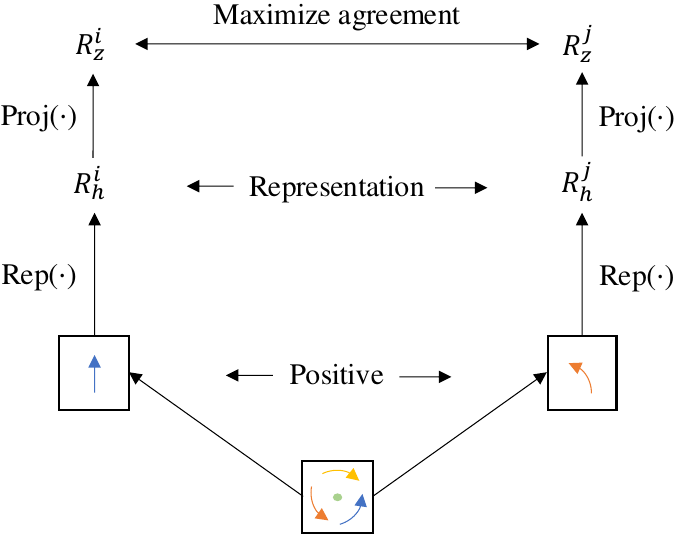}
    \caption{An illustration of the Intention Clusterer. Two trajectories are sampled as a positive pair based on predicted intention labels. A base encoder $\textbf{Rep}(\cdot)$ and a projection encoder $\textbf{Proj}(\cdot)$ are trained to maximize agreement using the contrastive clustering loss $\mathcal{L}_c$.}
    \label{fig:contrastive}
\end{figure}

We use projections of trajectories to construct positive and negative pairs based on predicted intention labels $\widehat{I}$ as illustrated in Figure~\ref{fig:contrastive}, and the detailed pairing scheme is shown in Figure~\ref{fig:model}. We define $R^i_{z^p, 1}$ and $R^i_{z^p, 2}$ are the first and the second projection for $i$th positive pair, and parallel notations $R^i_{z^n, 1}$ and $R^i_{z^n, 2}$ for $i$th negative pair. We randomly sample positive and negative pairs explicitly, resulting in $N_c$ examples. Given $i=1,2,\cdots,N_c$, We use $[R^i_{z^p}, R^i_{z^n}]=E^i$ to represented $i$th example $E^i$, where $(R^i_{z^p, 1}, R^i_{z^p, 2})=R^i_{z^p}$ and $(R^i_{z^n, 1}, R^i_{z^n, 2})=R^i_{z^n}$ denote $i$th positive and negative pair respectively. Then we compute the contrastive clustering loss $\mathcal{L}_c$ for examples set $E=\{E^1,E^2,\cdots,E^{N_c}\}$as:
\begin{equation}\label{eq20}
    \mathcal{L}_c=-\frac{1}{N_c}\sum^{N_c}_{i=1}\log\frac{\exp(\text{sim}(R^i_{z^p, 1}, R^i_{z^p, 2})/\mathcal{T})}{\sum_{(R^i_{z^\cdot, 1}, R^i_{z^\cdot, 2})\in E^i}\exp(\text{sim}(R^i_{z^\cdot, 1}, R^i_{z^\cdot, 2})/\mathcal{T})},
\end{equation}
where $\text{sim}(u, v)=u^{\top}\cdot v /\|u\|_2 \cdot \|v\|_2$ which denote the pairwise similarity and $\mathcal{T}$ denotes a hyper-parameter representing temperature.

By introducing the contrastive clustering loss $\mathcal{L}_c$, our \textit{Representation Learner} is optimized by jointly minimizing $\mathcal{L}_c$ and the classification loss $\mathcal{L}_r$. Compared with considering only classification loss $\mathcal{L}_r$, the model that also considers contrastive clustering loss $\mathcal{L}_c$ can pull the trajectories' representations of positive classes closer in the embedding space (e.g., straight and left), while pushing the trajectory representations of negative classes away from each other (e.g., static and straight). In addition, the representation of each trajectory (i.e., instance) will be more reflective of the fuzzy nature of road agents' intentions.

\subsection{Location Predictor}
\label{subsec:predictor}

\begin{figure}[t]
    \centering
    \includegraphics[width=\linewidth]{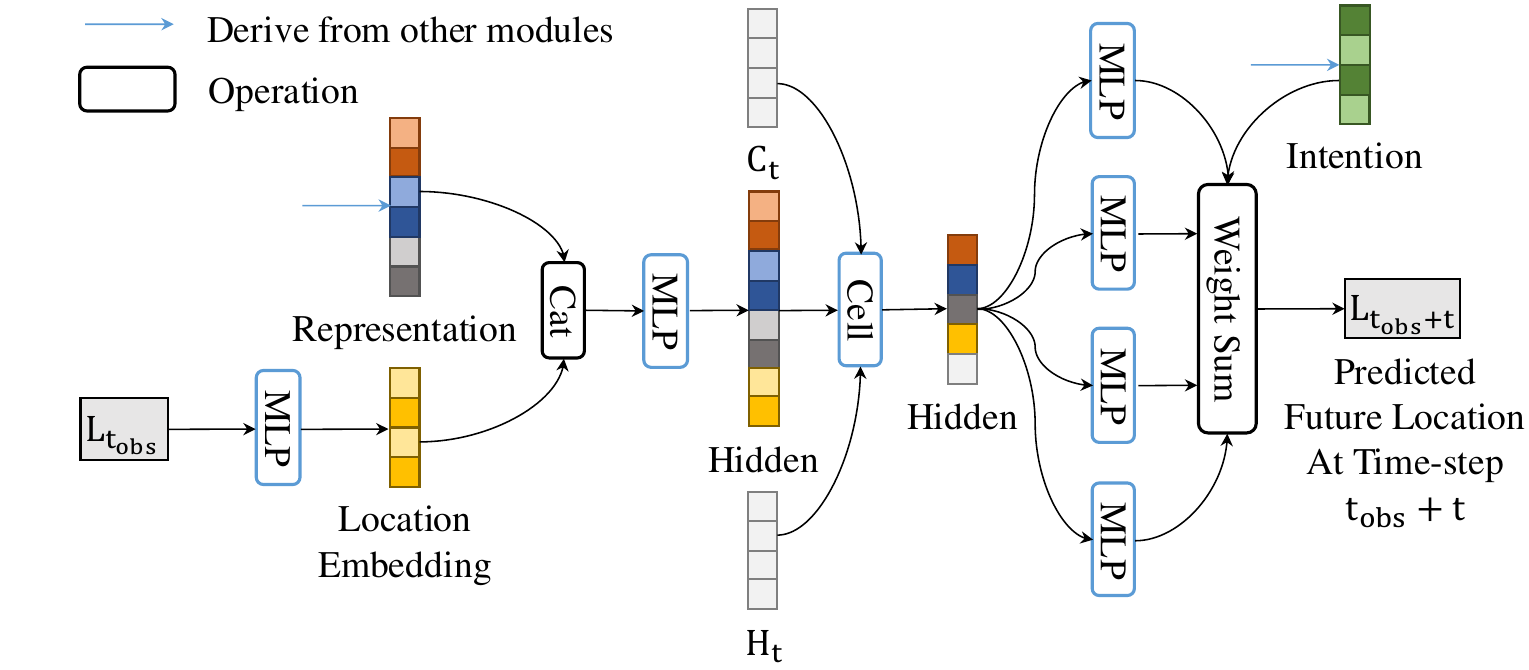}
    \caption{Structure of the \textit{Location Predictor}.}
    \label{fig:predictor}
\end{figure}

Most of the trajectory prediction methods leverage Recurrent Neural Network (\textsc{Rnn}) to encode and decode road agents' trajectories using multi-step mode for predictions \cite{zhang2019sr}. To be specific, the current time step's output is used as the input of the next time step. 
However, due to differences between the predicted and actual locations, the prediction error will inevitably accumulate as the time steps increase. 
In addition, the models of \textsc{Rnn} architectures are generally time-consuming, which has a significant impact on a real-time problem like trajectory prediction of autonomous vehicles in a road environment. Additionally, with all the information decoded and decoupled using the above-developed modules, there is no need to introduce the temporal module again for prediction.

To this end, we propose \textit{Location Predictor}, which is a time step-specific module that does not adopt an \textsc{Rnn} architecture and consists only of $\textbf{MLP}$s, and this substantially improves the training speed of our proposed $\textsc{Intent}$. We use notation $\mathcal{M}_{L_t}$ to denote the \textit{Location Predictor}, and it predicts the road agent's future location $\widehat{L_{t}}$ at time step $t \in \gamma_{pred}$ as described in section \ref{sec:method}. An overview of \textit{Location Predictor} is shown in Figure \ref{fig:predictor}, and it predicts a future location based on the road agent's intention $\widehat{I_S}$, extracted representation $R_h$ and its final observation location $L_{t_{obs}}$ of the observation trajectory. 

The latest observation location $L_{t_{obs}}$ is first encoded by a $\textbf{MLP}^{l}_{t}$ to obtain the location embedding of the road agent and it is further concatenated with the learned observation trajectory representation $R_h$. Then the $\textbf{MLP}^{c}_{t}$ is applied to generate the combined hidden vector $R_c$. The above process can be illustrated as:
\begin{equation}\label{eq21}
    R_c=\textbf{MLP}^{c}_{t}\left(\text{concat}\left(R_h, \tanh(\textbf{MLP}^{l}_{t}(L_{t_{obs}}))\right)\right),
\end{equation}
where $\tanh(\cdot)$ is the $tanh$ activation function.

After obtaining the combined feature $R_c$, we can predict location $\widehat{L_{t}}$ at time $t \in \gamma_{pred}$ as follows:
\begin{equation}\label{eq22}
    i_t, f_t, c_t, o_t = \text{split} ( (\textbf{W}^{i}_{t}\cdot {R}_{c} +\textbf{b}^{i}_{t})+(\textbf{W}^{h}_{t}\cdot {H}_{t}+\textbf{b}^{h}_{t}) ) ,
\end{equation}
\begin{equation}\label{eq23}
    h_t = \sigma(c_t)\cdot \tanh(\sigma(f_t)\cdot C_t+\sigma(i_t)\cdot\tanh(c_t)),
\end{equation}
\begin{equation}\label{eq24}
    \widehat{L_t} =\sum_{i=1}^{N}\widehat{I_S}[i] \cdot \textbf{MLP}^i_t(h_t),
\end{equation}
where $\sigma(\cdot)$ is the $sigmoid$ activation function, $H_t$ and $C_t$ are hidden vectors initialized to zeros, $\textbf{W}^i_t$, $\textbf{b}^i_t$, $\textbf{W}^h_t$, and $\textbf{b}^h_t$ are trainable parameters, and $i_t, f_t, c_t, o_t$ are different gates corresponding to \cite{hochreiter1997long}. $\widehat{I_S}[i]$ denotes the probability that the road agent's intention belong to intention class $i$, which is a scalar. $h_t$ is the hidden states. $t$ in Equation \ref{eq21} and \ref{eq24} is used to distinguish $\textbf{MLP}$s in different \textit{Location Predictor}s.

The $t_{pred}$ \textit{Location Predictor}s output $t_{pred}$ predicted future locations to form the predicted future trajectory $\widehat{Y}=\{\widehat{L}_{t_{obs}+1}, \cdots, \widehat{L}_{t_{obs}+t_{pred}}\}$. In addition to $\mathcal{L}_r$ and $\mathcal{L}_c$, we also apply MSE loss to measure the displacement between the predicted future trajectory $\widehat{Y}$ and the actual future trajectory $Y$, denoted as $\mathcal{L}_p$:
\begin{equation}\label{eq25}
    \mathcal{L}_p=\sum_{t=t_{obs}+1}^{t_{obs}+t_{pred}}\left \| L_t - \widehat{L}_t \right \|_{2}^{2}.
\end{equation}

\subsection{Training details}

We train $\textsc{Intent}$ end-to-end with a total loss function:
\begin{equation}\label{eq26}
    \mathcal{L}=\alpha\cdot\mathcal{L}_r+\beta\cdot\mathcal{L}_c+\mathcal{L}_p,
\end{equation}
where $\alpha$ and $\beta$ are hyper-parameters that trade off different loss terms. Additionally, we masked out those trajectories that are either  $max(\widehat{I_S})<Conf$ or without intention labels during training, and their loss is computed. Notation $Conf$ is adopted as the confidence lower bound to the soft intention label $\widehat{I_S}$.

\section{Experiments}
In this section, we present numerical experiments with both real-world trajectory datasets for pedestrian and autonomous vehicles.
\label{sec:exp}

\subsection{Datasets}\label{exp:dataset}
The datasets examined in the experiments are listed as follows.

\textbf{ETH \cite{pellegrini2010improving} \& UCY \cite{leal2014learning}} are two public pedestrian trajectory datasets with 1,536 pedestrians in 4 unique scenes, which contains 5 crowd sets including ETH-univ (eth), ETH-hotel (hotel), UCY-univ (univ), UCY-zara01 (zara1) and UCY-zara02 (zara2). We follow the leave-one-out data splitting settings in \cite{gupta2018social} and use $t_{obs}=8$ time steps (3.2 seconds)  for observation and predict $t_{pred}=12$ future time steps' (4.8 seconds).

\textbf{KITTI \cite{geiger2012we}} is an autonomous vehicle trajectory dataset. We use the version present in \cite{marchetti2020mantra}. KITTI contains 8,613 top-view trajectories for training and 2,509 trajectories for testing. We follow the setting in \cite{marchetti2020mantra} to set first 2s (20 time steps) as observation trajectories and predict future 1s-4s (10-40 time steps).

\textbf{SDD (Stanford Drone dataset) \cite{robicquet2016learning}} is a trajectory prediction benchmark from a top-down view, containing multiple moving road agents like humans, bicyclists, skateboarders, and vehicles. SDD captures 60 different scenes using drones at the campus of Stanford University. We use the dataset split defined in \cite{mangalam2020not}, and the time steps for observation and prediction are 8 (3.2s) and 12 (4.8s), respectively.

For all the above datasets, we only consider the trajectory information (2D coordinate sequence) and remove other environments (\textit{e.g.}, maps) and system (\textit{e.g.}, grouping, dynamics) information \cite{salzmann2020trajectron++} to ensure fair comparisons.

\renewcommand{\arraystretch}{1}
\begin{table}[h]
\caption{Hyper-parameters settings for \textsc{Intent}.} 
\centering 
\normalsize

\resizebox{\linewidth}{!}{
\begin{tabular}{l|ccccccc}

\toprule
 & \multicolumn{1}{c}{eth} & \multicolumn{1}{c}{hotel} & \multicolumn{1}{c}{univ} & \multicolumn{1}{c}{zara1} & \multicolumn{1}{c}{zara2} & \multicolumn{1}{c}{SDD} & \multicolumn{1}{c}{KITTI} \\  \cline{2-8} \hline 
learning rate                                              & le-3 & le-3 & 5e-3 & 5e-3 & 5e-3 & 2e-3 & 1e-3 \\
decay rate                                                 & 20\% & 20\% & 10\% & 10\% & 10\% & 10\% & 10\% \\
epochs per decay                                           & 10   & 20   & 30   & 20   & 20   & 50   & 50   \\
transformed                                                & Y    & Y    & N    & N    & N    & N    & N    \\
Conf                                                       & 0.99 & 0.98 & 0    & 0    & 0    & 0    & 0    \\
$\text{thresh}_{v / y}$                                    & 5  & 5  & 5  & 5 & 5 & 5 & 20 \\
$\text{V}_{a}$                                    & 0.01  & 0.01  & 0.01  & 0.01 & 0.01 & 0.01 & 0.01 \\
$\text{V}_{s}$                                    & 0.2  & 0.2  & 0.2  & 0.2 & 0.2 & 0.2 & 0.2 \\
$\text{V}_{lr}$                                    & 0.1  & 0.1  & 0.1  & 0.1 & 0.1 & 0.1 & 0.1 \\
$\alpha$                                                   & 0.8  & 1    & 0.2  & 2    & 1.8  & 1.8  & 0.2  \\
$\beta$                                                    & 1.2  & 1.8  & 1.2  & 1    & 1.8  & 0.5  & 0.8  \\
$\mathcal{T}$                                              & 0.1  & 0.1  & 0.1  & 0.07 & 0.06 & 0.06 & 0.09 \\
\bottomrule
\end{tabular}
}
\label{table:implementdetails}
\end{table}

\renewcommand{\arraystretch}{1}
\begin{table*}[h]
\caption{Comparisons of prediction accuracy for \textbf{deterministic} baselines (ADE / FDE in meters). Lower is better, \textbf{Bold} indicates the best, \underline{Underline} indicates the second best.} 
\centering 
\normalsize

\resizebox{0.85\textwidth}{!}{
\begin{tabular}{l|cccccc}

\toprule
 Model & \multicolumn{1}{c}{eth} & \multicolumn{1}{c}{hotel} & \multicolumn{1}{c}{univ} & \multicolumn{1}{c}{zara1} & \multicolumn{1}{c}{zara2} & \multicolumn{1}{c}{avg.} \\  \midrule
 \textsc{Social-Lstm} \cite{alahi2016social}             & 1.09 / 2.35 & 0.79 / 1.76 & 0.67 / 1.40 & 0.47 / 1.00 & 0.56 / 1.17 & 0.72 / 1.54 \\
 \textsc{Social-Gan-1} \cite{gupta2018social}            & 1.13 / 2.21 & 1.01 / 2.18 & 0.60 / 1.28 & \textbf{0.42} / 0.91 & 0.52 / 1.11 & 0.74 / 1.54 \\
 \textsc{Social-Attention} \cite{vemula2018social}       & 1.39 / 2.39 & 2.51 / 2.91 & 0.88 / 1.75 & 1.25 / 2.54 & 1.01 / 2.17 & 1.41 / 2.35 \\
 \textsc{Cidnn} \cite{xu2018encoding}                    & 1.25 / 2.32 & 1.31 / 1.86 & \underline{0.53} / 1.14 & 0.90 / 1.28 & 0.50 / 1.04 & 0.89 / 1.73 \\
 \textsc{TrafficPredict} \cite{ma2019trafficpredict}     & 5.46 / 9.73 & 2.55 / 3.57 & 3.31 / 6.37 & 4.32 / 8.00 & 3.76 / 7.20 & 3.88 / 6.97 \\
 \textsc{Matf} \cite{zhao2019multi}                      & 1.33 / 2.49 & 0.51 / 0.95 & 0.56 / 1.19 & 0.44 / 0.93 & \underline{0.34} / \underline{0.73} & 0.64 / 1.26 \\
 \textsc{Pitf} \cite{liang2019peeking}                   & 0.88 / 1.98 & \underline{0.36} / 0.74 & 0.62 / 1.32 & \underline{0.42} / 0.90 & \underline{0.34} / 0.75 & 0.52 / 1.14 \\
\textsc{Stgat-1} \cite{huang2019stgat}                  & 1.03 / 2.20 & 0.59 / 1.21 & \underline{0.53} / 1.17 & 0.43 / 0.94 & 0.68 / 1.49 & 0.65 / 1.40 \\
 \textsc{FvTraj} \cite{bi2020can}                        & \textbf{0.62} / \textbf{1.23} & 0.53 / 1.10 & 0.57 / 1.19 & \underline{0.42} / \underline{0.89} & 0.38 / 0.79 & \underline{0.50} / \underline{1.04} \\
 \textsc{Social-Stgcnn-1} \cite{mohamed2020social}       & 1.01 / 1.83 & 0.74 / 1.42 & 0.57 / \underline{1.13} & 0.51 / 0.97 & 0.71 / 1.38 & 0.71 / 1.35 \\
 \textsc{Sti-Gan-1} \cite{huang2021sti}                  & 0.94 / 1.81 & 0.82 / 1.28 & 0.57 / 1.23 & 0.45 / 0.96 & 0.40 / 0.87 & 0.64 / 1.23 \\
 \textsc{Tf} \cite{giuliari2021transformer}              & 1.03 / 2.10 & \underline{0.36} / \underline{0.71} & \underline{0.53} / 1.32 & 0.44 / 1.00 & \underline{0.34} / 0.76 & 0.54 / 1.17 \\
\midrule
 \textsc{Intent} (ours)                                         & \underline{0.86} / \underline{1.62} & \textbf{0.24} / \textbf{0.44} & \textbf{0.51} / \textbf{1.07} & \textbf{0.42} / \textbf{0.88} & \textbf{0.33} / \textbf{0.71} & \textbf{0.48} / \textbf{0.96} \\ \bottomrule

\end{tabular}
}
\label{table:deterministic}
\end{table*}

\renewcommand{\arraystretch}{1}
\begin{table*}[h]
\caption{Prediction accuracy comparison of stochastic baselines (ADE / FDE in meters). Lower is better, \textbf{Bold} indicates the best, \underline{Underline} indicates the second best.} 
\centering 
\normalsize

\resizebox{0.85\textwidth}{!}{
\begin{tabular}{l|cccccc}

\toprule
Stochastic & \multicolumn{1}{c}{eth} & \multicolumn{1}{c}{hotel} & \multicolumn{1}{c}{univ} & \multicolumn{1}{c}{zara1} & \multicolumn{1}{c}{zara2} & \multicolumn{1}{c}{avg.} \\  \midrule
\textsc{Social-Gan} \cite{gupta2018social}       & 0.87 / 1.49 & 0.67 / 1.37 & 0.76 / 1.52 & 0.35 / 0.68 & 0.42 / 0.84 & 0.61 / 1.21 \\
\textsc{Social-Gan-p} \cite{sadeghian2019sophie} & 0.87 / 1.62 & 0.67 / 1.37 & 0.76 / 1.52 & 0.35 / 0.68 & 0.42 / 0.84 & 0.61 / 1.21 \\
\textsc{Cgns} \cite{li2019conditional}           & \textbf{0.62} / \underline{1.40} & 0.70 / 0.93 & \textbf{0.48} / 1.22 & 0.32 / \textbf{0.59} & \underline{0.35} / \underline{0.71} & \underline{0.49} / \underline{0.97} \\ 
\textsc{SoPhie} \cite{sadeghian2019sophie}       & 0.70 / 1.43 & 0.76 / 1.67 & 0.54 / 1.24 & \underline{0.30} / 0.63 & 0.38 / 0.78 & 0.54 / 1.15 \\
\textsc{Gat} \cite{kosaraju2019social}           & \underline{0.68} / \textbf{1.29} & 0.68 / 1.40 & 0.57 / 1.29 & \textbf{0.29} / \underline{0.60} & 0.37 / 0.75 & 0.52 / 1.07 \\
\textsc{Social-Bigat} \cite{kosaraju2019social}  & 0.69 / \textbf{1.29} & \underline{0.49} / 1.01 & 0.55 / 1.32 & \underline{0.30} / 0.62 & 0.36 / 0.75 & \textbf{0.48} / 1.00 \\
\textsc{Sti-Gan} \cite{huang2021sti}             & 0.77 / 1.53 & 0.70 / \underline{0.73} & \underline{0.53} / \underline{1.20} & 0.33 / \textbf{0.66} & \textbf{0.33} / \textbf{0.66} & 0.53 / \textbf{0.96} \\
\midrule

\textsc{Intent} (Deterministic) & 0.86 / 1.62 & \textbf{0.24} / \textbf{0.44} & \underline{0.53} / \textbf{1.15} & 0.42 / 0.88 & \textbf{0.33} / \underline{0.71} & \textbf{0.48} / \textbf{0.96} \\ \bottomrule

\end{tabular}
}
\label{table:stochastic}
\end{table*}

\subsection{Implementation Details \& Reproducibility}\label{exp:implement}
All the experiments are conducted on a desktop with Intel Core i9-10990K CPU @3.7GHz $\times$ 10, 2666MHz $\times$ 2 $\times$ 16GB RAM, GeForce RTX 3080 $\times$ 2, 500GB SSD. We implement $\textsc{Intent}$ based on Pytorch \cite{paszke2019pytorch}. Parameters of $\textsc{Intent}$ are randomly initialized and optimized by Adam optimizer \cite{kingma2014adam}. 
For hyper-parameters tuning, we conduct heuristic search based on Optuna framework \cite{akiba2019optuna} by exploring $\alpha \in \{$0.2, 0.5, 0.8, 1, 1.2, 1.5, 1.8, 2$\}$, $\beta \in \{$0.2, 0.5, 0.8, 1, 1.2, 1.5, 1.8, 2$\}$ and $\mathcal{T} \in \{$0.05, 0.06, 0.07, 0.08, 0.09, 0.1$\}$. For univ dataset, we perform data augmentation of the training set by randomly rotating and translating to solve the issue of low training data. Dimensions of spatial embeddings are set to 32 and 64 for hidden layers. Detailed hyper-parameters settings are summarized in Table \ref{table:implementdetails}.

\subsection{Evaluation Metrics}\label{exp:metric}
In this paper, we first adopted two commonly used \cite{mangalam2020not,gupta2018social,sadeghian2019sophie} metrics \textit{ADE} and \textit{FDE} for trajectory prediction evaluation. In addition, we measured the training and inference \textit{TIME} to demonstrate our model's high efficiency and the benefits of its lightweight design.
\begin{itemize}[leftmargin=*]
    \item \textit{Average Displacement Error (ADE)}: Mean $\ell_{2}$ distance between actual and predicted future trajecotries.
    \item \textit{Final Displacement Error (FDE)}: $\ell_{2}$ distance between final locations of actual and predicted future trajectories.
    \item \textit{TIME}: Training and inference times of different models.
\end{itemize}

\subsection{Main Results}

In the section, we discuss the performance of \textsc{Intent} in different datasets separately by comparing with other baseline methods.

\textbf{ETH-UCY dataset}. We compare $\textsc{Intent}$ with a wide range of baselines, including deterministic and stochastic methods. Quantitative results reflect the superiority of \textsc{Intent} in prediction accuracy and training efficiency. We consider stochastic methods that generate only one sample as deterministic methods (with ``-1'' as suffix). Table \ref{table:deterministic} shows the quantitative results of our \textsc{Intent} against state-of-the-art deterministic methods. \textsc{Intent} outperforms all deterministic baselines on average. In contrast to multiple baseline methods that emphasize interaction, dynamics and context information \cite{gupta2018social,sadeghian2019sophie}, our \textsc{Intent}, which only learns from past trajectory information, obtains 4.0\% and 7.7\% improvements in \textit{ADE} and \textit{FDE} metrics comparing to the second-best baseline on average.

In addition to comparisons with deterministic methods, \textsc{Intent} even achieves better prediction accuracy when compared with some stochastic methods. Given the number of samples (\textit{e.g.}, 20), these sampling-based stochastic methods have the probability to sample and generate multiple trajectories with diverse shapes and lengths, while they only report the error of the one generated trajectory that is closest to the actual trajectory (Ground Truth). As a result, stochastic methods often achieve more accurate predictions than deterministic methods. We still compare with stochastic methods that sample 20 times and report the best prediction result of 20 predictions. Table \ref{table:stochastic} shows that overall baseline prediction accuracy increased drastically compared with the result in Table \ref{table:deterministic}. Our \textsc{Intent} still has the highest average prediction result across the five scenarios.

\renewcommand{\arraystretch}{1.02}
\begin{table}[h]
\caption{Efficiency comparisons in both training and inference settings with representative baseline methods.} 
\centering 
\normalsize

\resizebox{\linewidth}{!}{
\begin{tabular}{l|l|cccccc}

\toprule
Time                       & Models & eth            & hotel          & univ           & zara1          & zara2          & avg.           \\ \midrule
\multirow{3}{*}{Training (s)}     & \textsc{TF} \cite{giuliari2021transformer}     & 3979.2         & 4254.7         & 9188.4         & 4380.9         & 5386.4         & 5437.92        \\
                           & \textsc{Lstm} \cite{hochreiter1997long}   & 392.1          & 653.7          & 598.9          & 567.9          & 393.9          & 521.3          \\
                           & \textsc{Intent} & \textbf{133.3} & \textbf{190.7} & \textbf{480.4} & \textbf{335.9} & \textbf{247.4} & \textbf{277.5} \\ \midrule
\multirow{3}{*}{Inference (ms)} & \textsc{TF} \cite{giuliari2021transformer}     & 0.453         & 0.378         & 0.416         & 0.384         & 0.401         & 0.406        \\
                           & \textsc{Lstm} \cite{hochreiter1997long}   & 0.065          & \textbf{0.038}          & 0.008          & 0.045          &  \textbf{0.017}          & 0.035          \\
                           & \textsc{Intent} & \textbf{0.032} & 0.042 & \textbf{0.006} & \textbf{0.035} & 0.021 &\textbf{0.027} \\ \bottomrule
\end{tabular}
}
\label{table:time}
\end{table}

Moreover, to demonstrate the lightweight nature of \textsc{Intent}, we selected two baseline models representing RNN-based and Transformer-based architectures, as other baselines usually build upon these with additional components. We compared the training time of \textsc{Intent} with that of the Recurrent Neural Network (RNN)-based model \textsc{Lstm} \cite{hochreiter1997long} and the Transformer-based model \textsc{Tf} \cite{giuliari2021transformer}. The numerical training time results are shown in Table \ref{table:time}. \textsc{Intent} reduces training time by 94.90\% and 46.77\% compared to \textsc{Tf} and \textsc{Lstm}, respectively. 
These results demonstrate the high efficiency achieved with the lightweight MLP-based design of \textsc{Intent}. 
Additionally, inference time plays a pivotal role in real-world deployment, so we also evaluated it to provide a comprehensive assessment of the model's efficiency. As presented in Table \ref{table:time}, show that \textsc{Intent} outperforms both \textsc{Tf} and \textsc{Lstm}, making it a more suitable choice for real-time applications where quick response times are critical. This substantial improvement in both training and inference times underscores the practicality and effectiveness of using MLP-based architectures for trajectory prediction tasks, particularly in scenarios requiring high efficiency and low latency.

\renewcommand{\arraystretch}{1.5}
\begin{table}[h]
\caption{Prediction accuracy comparisons on Standord Drone Dataset (ADE / FDE). \textbf{Bold} indicates the best, \underline{Underline} indicates the second best.} 
\centering 

\resizebox{\linewidth}{!}{
\begin{tabular}{cccc|c}

\toprule
\textsc{Social-Lstm} & \textsc{Social-Gan} & \textsc{Multiverse-1} & \textsc{PECNet-1} & \textsc{Intent} (ours)
\\ \midrule
31.19 / 56.79 & 27.23 / \underline{41.44} & \underline{21.9} / 42.8 & 30.31 / 65.13 & \textbf{19.56} / \textbf{39.09} \\
\bottomrule
\end{tabular}
}
\label{table:sdd}
\end{table}

\textbf{Stanford Drone dataset}. Our work aims to understand the information contained in trajectories to make more accurate trajectory predictions, so \textsc{Intent} should be powerful at predicting future trajectories regardless of types of road agents. Table \ref{table:sdd} shows the results of \textsc{Intent} against different baselines on the SDD dataset. Since most recent works only report stochastic results, we consider single sample results for fair comparisons. \textsc{Intent} has a 10.7\% and 5.7\% improvement on SDD dataset over the best baseline on \textit{ADE} and \textit{FDE}, respectively.

\renewcommand{\arraystretch}{1.1}
\begin{table}[h]
\caption{Prediction accuracy comparisons on KITTI dataset (ADE / FDE). \textbf{Bold} indicates the best, \underline{Underline} indicates the second best.} 
\centering 

\resizebox{\linewidth}{!}{
\begin{tabular}{l|cccc}

\toprule
\multirow{1}{*}{KITTI} & \multicolumn{1}{c}{1s} & \multicolumn{1}{c}{2s} & \multicolumn{1}{c}{3s} & \multicolumn{1}{c}{4s} \\  \midrule
\textsc{Kalman} \cite{marchetti2020mantra}             & 0.51 / 0.97 & 1.14 / 2.54 & 1.99 / 4.71 & 3.03 / 7.41 \\
\textsc{Linear} \cite{marchetti2020mantra}             & \underline{0.20} / \underline{0.40} & \underline{0.49} / \underline{1.18} & \underline{0.96} / \underline{2.56} & \underline{1.54} / 4.73 \\
\textsc{Const-Vel} \cite{scholler2020constant}         & 0.34 / 0.56 & 0.85 / 1.79 & 1.60 / 3.72 & 2.55 / 6.24 \\
\textsc{Social-Lstm} \cite{alahi2016social}            & 0.53 / 1.07 & 1.05 / 2.10 & 1.93 / 3.26 & 2.91 / 5.47 \\
\textsc{Social-Gan} \cite{gupta2018social}             & 0.29 / 0.43 & 0.67 / 1.34 & 1.26 / 2.94 & 2.07 / 5.22 \\
\textsc{Gated-Rn} \cite{choi2019looking}               & 0.34 / 0.62 & 0.70 / 1.72 & 1.30 / 3.34 & 2.09 / 5.55 \\
\textsc{Desire-1} \cite{lee2017desire}                 &  \text{ }\text{ }--\text{ }\text{ }  / 0.51 &  \text{ }\text{ }--\text{ }\text{ }  / 1.44 & \text{ }\text{ }--\text{ }\text{ }    / 2.76 & \text{ }\text{ }--\text{ }\text{ }  / \underline{4.45} \\
\textsc{Mantra-1} \cite{marchetti2020mantra}           & 0.24 / 0.44 & 0.57 / 1.34 & 1.08 / 2.79 & 1.78 / 4.83 \\
\midrule
\textsc{Intent} (ours)                                        & \textbf{0.17} / \textbf{0.34} & \textbf{0.42} / \textbf{0.99} & \textbf{0.85} / \textbf{2.21} & \textbf{1.46} / \textbf{3.85} \\
\bottomrule

\end{tabular}
}
\label{table:kitti}
\end{table}

\textbf{KITTI dataset}. Table \ref{table:kitti} shows the results on the KITTI dataset, which contains trajectories of autonomous vehicles. 
Vehicles' trajectories are generally less complex and smoother than human trajectories, which explain the superiority performances of linear regressor model to some extent. The prediction results reported in Table \ref{table:kitti} proves that \textsc{Intent} is robustly powerful in different time spans, and results on both \textit{ADE} and \textit{FDE} demonstrate the state-of-the-art results for all time spans (1s-4s). \textsc{Intent} yields 15\%, 14.3\%, 11.5\%, 5.2\% improvements on \textit{ADE} and 15\%, 16.1\%, 17.6\%, 13.5\% improvements on \textit{FDE} compared with the best baseline, for 1s, 2s, 3s, and 4s, respectively.

\begin{figure}[h]
    \centering
    \includegraphics[width=\linewidth]{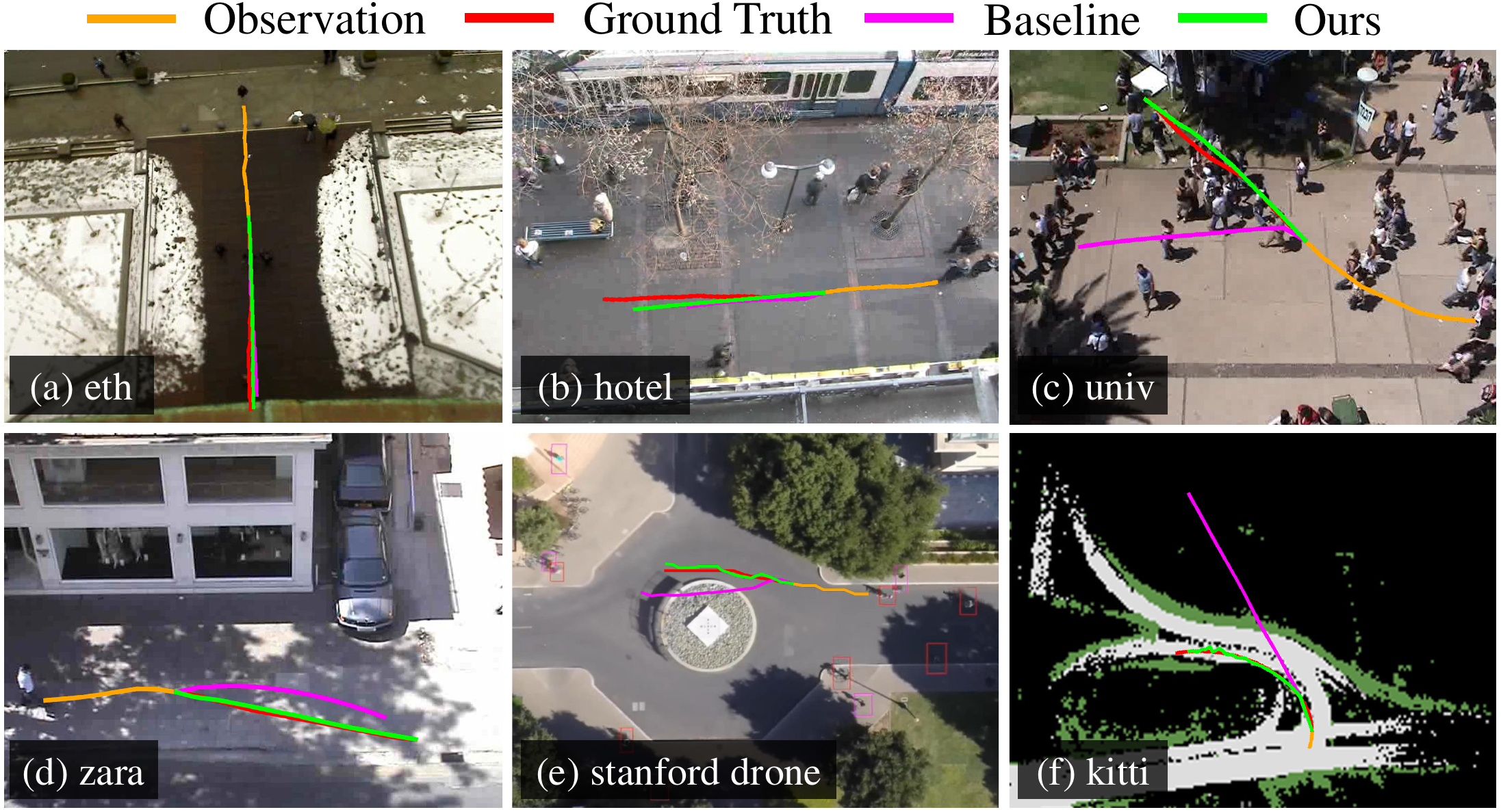}
    \caption{Visualization of predictions by \textsc{Intent}.}
    \label{fig:v_trajs}
\end{figure}

\textbf{Qualitative evaluations.} Figure \ref{fig:v_trajs} shows qualitative comparisons between \textsc{Intent} and baseline method \textsc{Lstm}. The baseline method could be misled by the limited observation locations and generate poor predictions as shown in Figure \ref{fig:v_trajs}(c), (f).  In contrast, by taking advantages of predicted intentions, \textsc{Intent} generates more accurate and reasonable predictions that are closer to the ground true trajectories.

\begin{figure}[h]
    \centering
    \includegraphics[width=\linewidth]{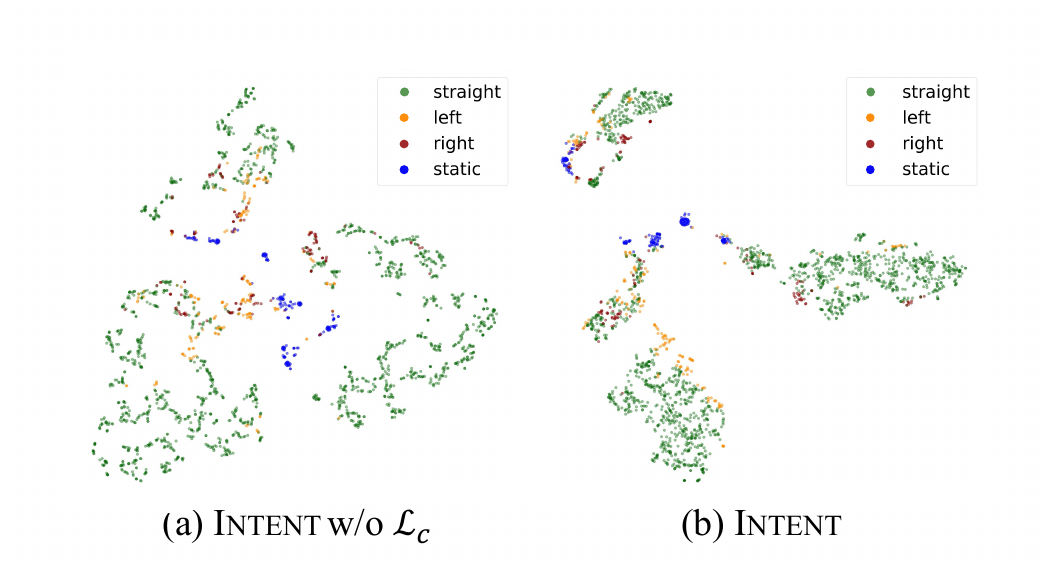}
    \caption{Trajectories representations of zara1 test set visualized by t-SNE \cite{van2013barnes} in $\mathbb{R}^2$.}
    \label{fig:tsne}
\end{figure}

\subsection{Analysis of Interpretability}

\begin{figure*}[t]
    \centering
    \includegraphics[width=\textwidth]{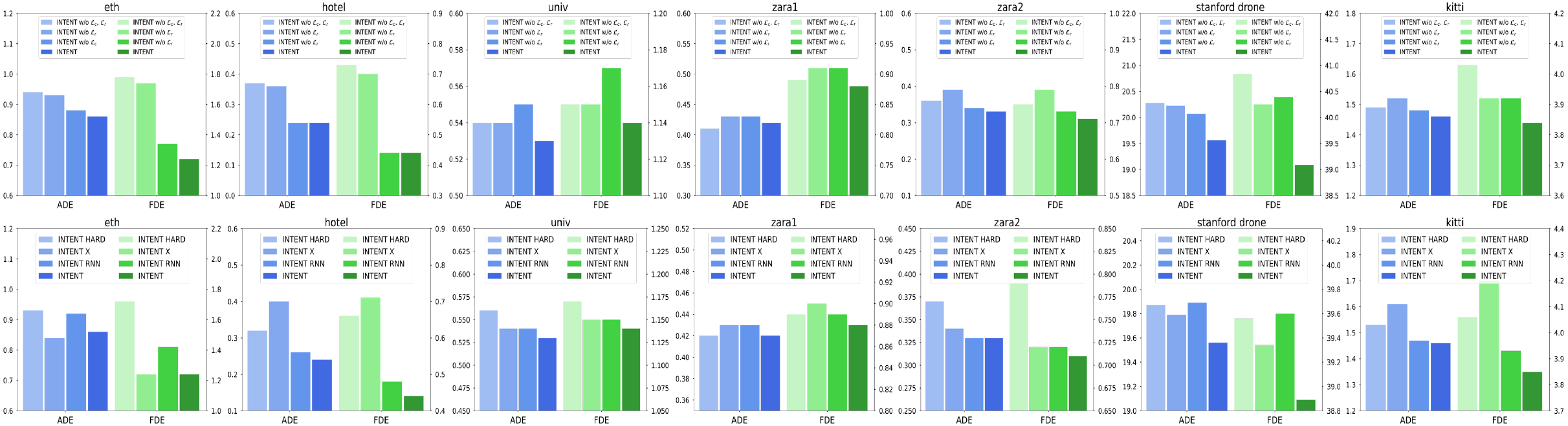}
    \caption{Ablation study comparing \textsc{Intent} with six variants: (1) \textsc{Intent} w/o $\mathcal{L}_c$, $\mathcal{L}_r$, (2) \textsc{Intent} w/o $\mathcal{L}_c$, (3) \textsc{Intent} w/o $\mathcal{L}_r$, (4) \textsc{Intent} HARD, (5) \textsc{Intent}-$X$, and (6) \textsc{Intent} RNN. All variants use the same settings as \textsc{Intent}.}
    \label{fig:ablations}
\end{figure*}

\begin{figure}[h]
    \centering
    \includegraphics[width=\linewidth]{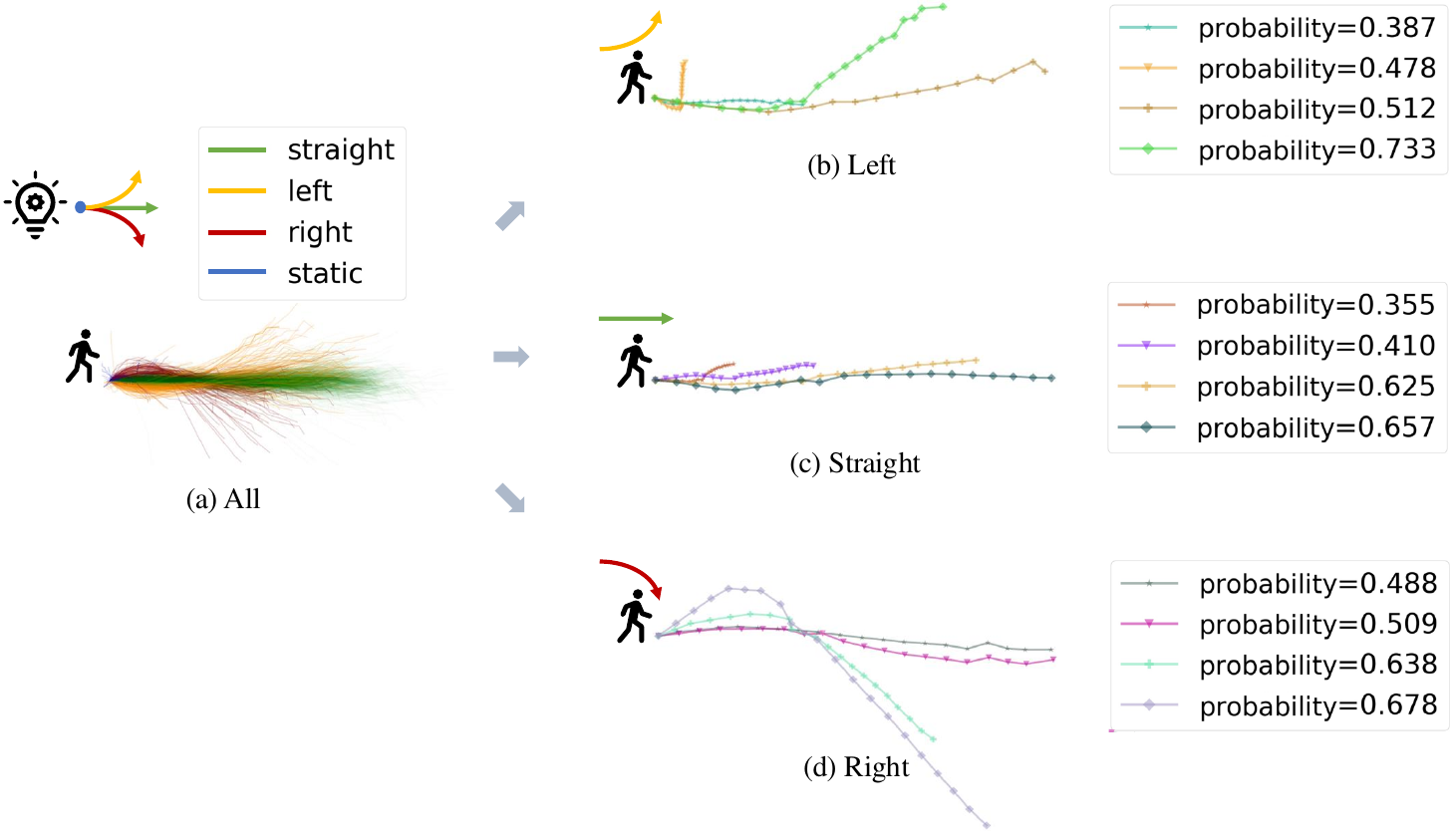}
    \caption{Visualization of transformed trajectories with corresponding probabilities for different intention classes.}
    \label{fig:visual}
\end{figure}

To analyze the potential interpretability of \textsc{Intent}, we visualize the trajectory and the corresponding probability associated with each of the intention class. To this end, we visualized predicted trajectories of the zara1 dataset for different intention classes in Figure \ref{fig:visual}.

From Figure \ref{fig:visual}(a), we can see that the classification of the intention class is accurate and intuitive. Trajectory also reflects the feature of clustering by intention classes. Moreover, we have selected trajectories with different probability from each predicted intention class for visualization, and trajectories that predicted belong to left, straight, and right are plotted in (b), (c), and (d) of Figure \ref{fig:visual}, respectively. For the left intention class / the right intention class, trajectories are more likely to bend up / bend down while probabilities increase. Furthermore, trajectories belonging to the straight intention class will be smoother with the increasing probability. The above observations demonstrate that, to some extent, the intentions of road agents can be modeled by dividing them into four intention classes, which also prove the effectiveness of \textit{Observation Feature Extractor} and \textit{Representation Learner}.

\subsection{Ablation Study}

We also conduct ablation study on \textsc{Intent} and results of its variants are shown in Figure \ref{fig:ablations}. We verify six variants of \textsc{Intent}: (1) \textsc{Intent} w/o $\mathcal{L}_r, \mathcal{L}_c$, (2) \textsc{Intent} w/o $\mathcal{L}_r$, (3) \textsc{Intent} w/o $\mathcal{L}_c$.  (4) \textsc{Intent} Hard. (5) \textsc{Intent}-$X$. (6) \textsc{Intent} \textsc{Rnn}. The first row of Figure \ref{fig:ablations} display that \textsc{Intent} w/o $\mathcal{L}_r, \mathcal{L}_c$'s prediction error are the worst in most cases, this demonstrates the clustered and diverse nature of the trajectory. Improvement ratios gained by adding $\mathcal{L}_r$ and $\mathcal{L}_c$ may vary depending on the change of the dataset. However, the prediction accuracy improves more after adding $\mathcal{L}_r$ in most cases, this demonstrate the effectiveness of using intention classes to model road agents' intentions. In particular, the results of our \textsc{Intent} reach optimality, which also confirms that the trajectory representation learned under the guidance of both $\mathcal{L}_r$ and $\mathcal{L}_c$ is optimal. The second row of Figure \ref{fig:ablations} compare \textsc{Intent} with last three variants. \textsc{Intent} HARD decodes hidden states $h_t$ in Equation \ref{eq23} with a single \textbf{MLP} which corresponds to the predicted intention class. \textsc{Intent}-$X$ is the variant that directly use location coordinates without \textit{Observation Feature Extractor}.

To understand how the trajectories representations change by minimizing loss $\mathcal{L}_r$ and $\mathcal{L}_c$, we visualize the extracted representations distribution of the zara1 test set in $\mathbb{R}^2$ using t-SNE \cite{van2013barnes}. As shown in Figure \ref{fig:tsne}(a), representations of \textsc{Intent w/o $\mathcal{L}_c$} exhibits the clustering features corresponding to intention classes. 
Figure \ref{fig:tsne}(b) shows representations distribution of \textsc{Intent}. The learned representations are more clustered by minimizing $\mathcal{L}_c$, and representations of similar trajectories that belong to different intention classes may group based on our positive and negative pairs definition. Representations that belong to the left intention class are more isolated from representations that belong to the right intention class at the same time.

\begin{figure}[h]
    \centering
    \includegraphics[width=\linewidth]{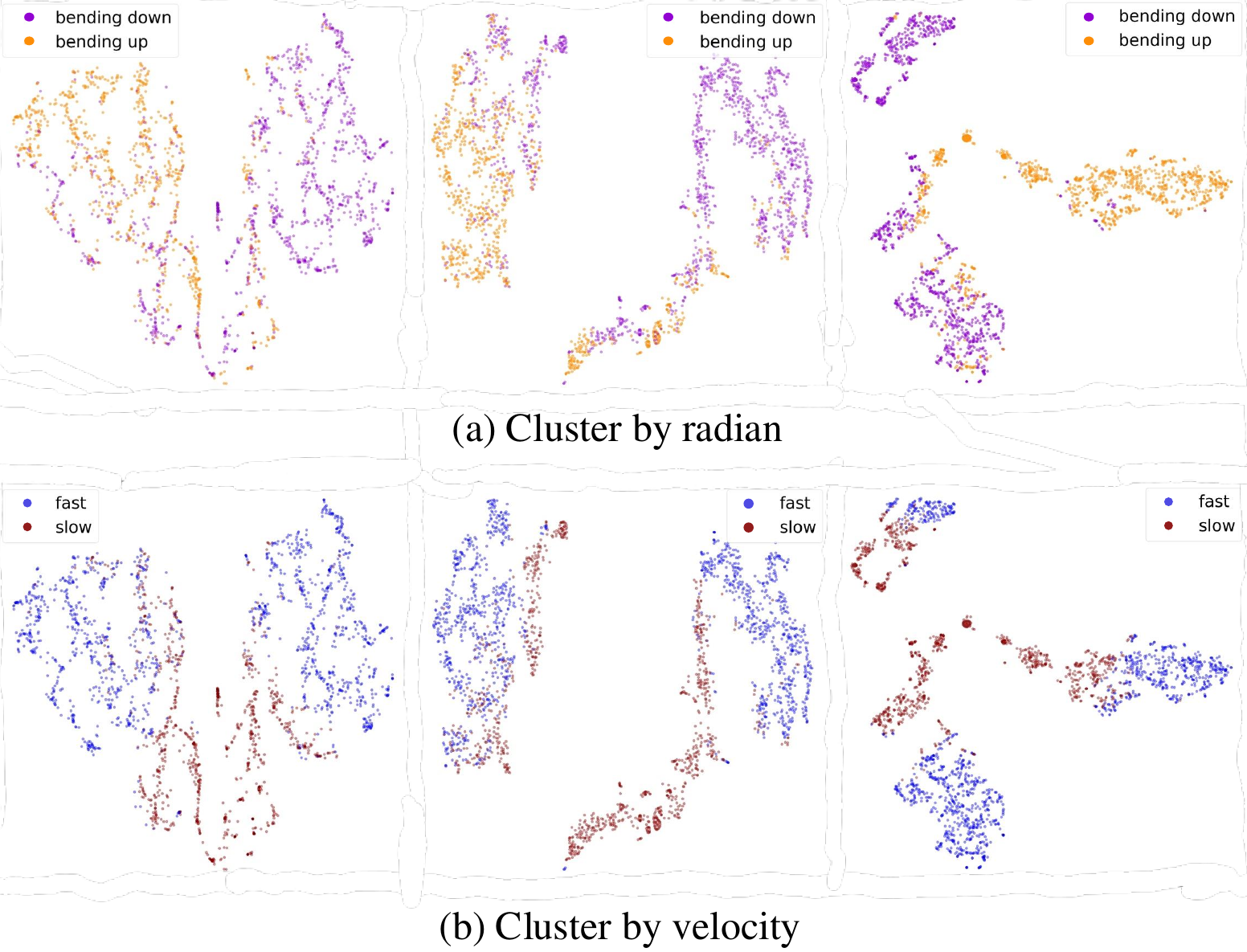}
    \caption{Visualization of the trajectory representation distribution of different ablation methods. Variants from left to right: \textsc{Intent}-$X$ w/o intentions, \textsc{Intent}-$X$, \textsc{Intent}}
    \label{fig:tsnevr}
\end{figure}

To analyze how road agents' intentions contribute to the trajectories representation learning and verify the effectiveness of adopting observation features as input instead of location coordinates $X$, we also visualize the learned trajectories representations distribution of two variants of \textsc{Intent} using t-SNE. As shown in Figure \ref{fig:tsnevr}, (a) demonstrates the visualizations of \textsc{Intent}-$X$ w/o intentions, this method replace input of \textsc{Intent} by observation trajectory $X$ and remove all operation that relevant to intentions (which can be denoted as \textsc{Intent}-$X$ w/o $\mathcal{L}_r$, $\mathcal{L}_c$). (b) differs from (a) in that it retains the operations associated with intentions. (c) displays trajectory representation distribution that is clustered by different observation features (note that trajectory representations distribution is the same as Figure \ref{fig:tsne}(b)'s representations distribution).

The distribution of representations in (a) is more uniform compared to representations in (b) and (c), and some representations in different clusters are mixed. By considering intentions, trajectories representations in (b) have more clustered distribution and greater separation between samples classified differently by speed and bending than in (a). However, the boundaries between clusters are still vague. Trajectory representations distribution in (c) demonstrates the effectiveness of inputting observation features instead of observation locations $X$. In comparison with (a) and (b), trajectories in (c) indicate more clear clustering characteristics of the trajectories representations distribution. At the same time, the separation between different clustered distributions is also clearer, which to some extent explains the superior performance of \textsc{Intent}.

\subsection{Analysis of Parameter Sensitivity}

\begin{figure}[h]
    \centering
    \includegraphics[width=.49\textwidth]{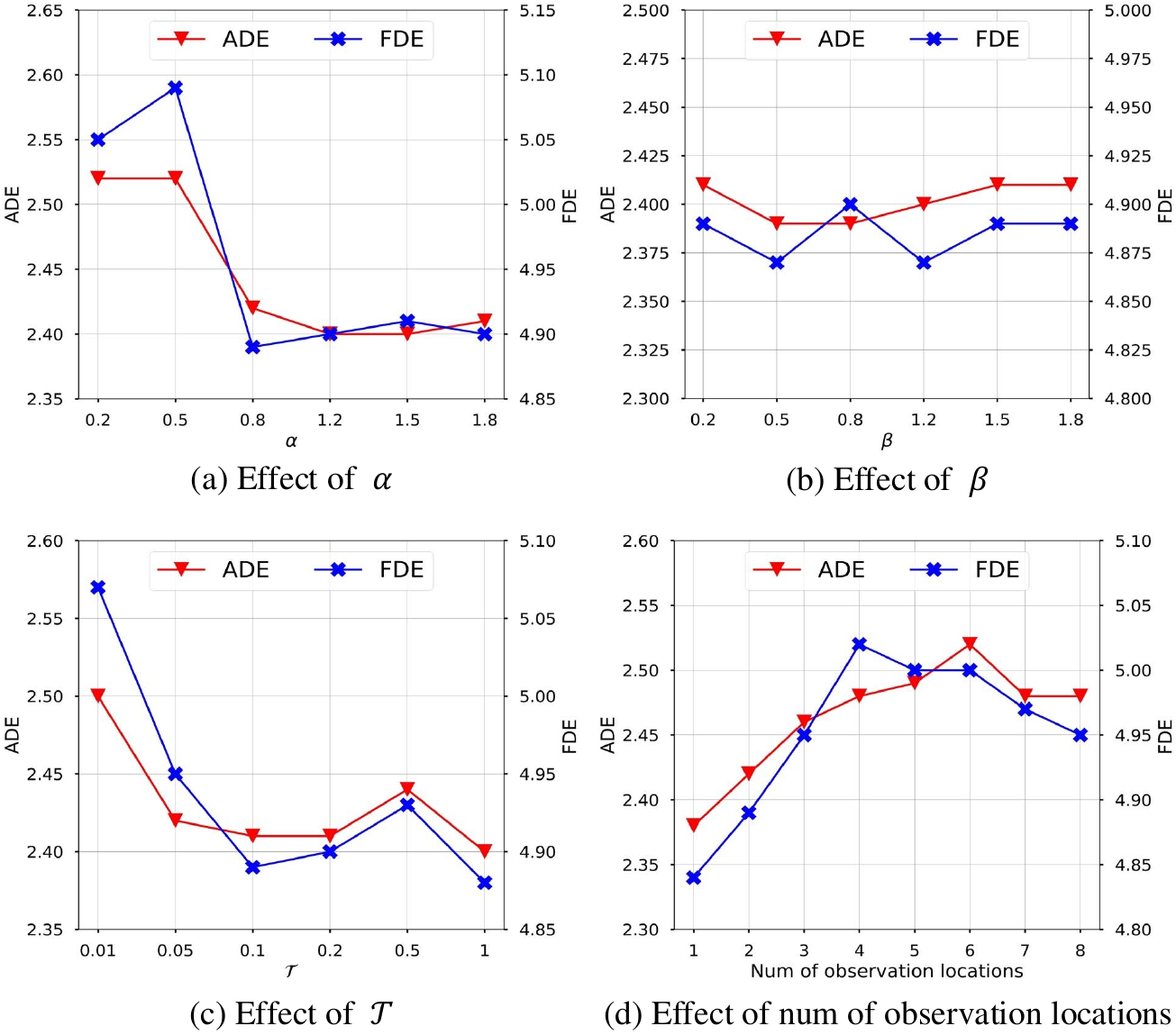}
    \caption{Parameter sensitivity on ETH-UCY dataset.}
    \label{fig:para}
\end{figure}
In this section, we conduct the parameter sensitivity experiments on the ETH-UCY dataset by varying the hyper-parameters and a summary of \textit{ADE} and \textit{FDE} over all datasets is generated.

\textbf{Effect of $\alpha$}. We first vary the parameter $\alpha$ from $0.2$ to $1.8$ and the results are shown in Figure \ref{fig:para}(a). The performance of \textsc{Intent} first improves from 0.2 to 0.8 and then degrades slightly from 1 to 1.8. This observation proves the effectiveness of intention labels. A probable reason for performance decay is that too large $\alpha$ makes our model focus on classification accuracy and may harm the trajectory prediction task.

\textbf{Effect of $\beta$}. We also evaluate the sensitivity of the parameter $\beta$, which affects the extent of similarities between trajectories (positive pairs) that belong to different intention classes. As shown in Figure \ref{fig:para}(b), prediction results have some oscillations in \textit{FDE} metric and the variation trend of \textit{ADE} results may be caused by the same reason mentioned in a discussion of the effect of the parameter $\alpha$.

\textbf{Effect of $\mathcal{T}$}. \cite{wang2021understanding} states that a small temperature tends to generate a more uniform distribution, which is verified in Figure \ref{fig:tsne}. 
We tune the parameter temperature $\mathcal{T}$ from 0.01 to 1, and the prediction accuracy is shown in Figure \ref{fig:para}(c). Both \textit{ADE} and \textit{FDE} decrease and then oscillate. The possible reason is when $\mathcal{T}$ is small, the uniform distribution may lack representation capability. When $\mathcal{T}$ gradually increases, representations of trajectories may be more distinguishable and hence minimize the $\mathcal{L}_r$ and $\mathcal{L}_c$. The oscillation might be due to the instability of  adversarial  learning.

\textbf{Effect of the number of observation locations}. Finally, we evaluate the impact of the number of input observation locations. Results are reported in Figure \ref{fig:para}(d). 
The optimal observation length exists, and hence the cross-validation for choosing the optimal length is essential for obtaining the powerful representation of trajectories.

\section{Conclusion}
\label{sec:conclusion}

We propose a novel framework \textsc{Intent} for predicting future trajectories of road agents based on their intentions. Our experimental results show that the proposed contrastive clustering could capture road agents' intentions with interpretability and improve prediction accuracy. Additionally, without the \textsc{Rnn} structure or attention mechanism, our model can still achieve good trajectory prediction with low computational complexity. 
We hope that our work could contribute to the development of trajectory prediction methods by emphasizing the necessity for understanding and modeling intentions, as well as triggering in-depth studies for mining information contained in the trajectories of road agents.

In terms of future studies, it would be interesting to particularly estimate the intentions of those vulnerable road agents (e.g.,  pedestrians with disabilities, autonomous vehicles carrying emergency patients). With the predicted trajectories, it would be crucial to quantify the potential risks of collision with different road agents. Methodology-wise, agent-based contrastive learning modules can be developed to differentiate the behavior (in addition to intentions) of different types of road agents, and the proposed \textsc{Intent} can be further extended to predict stochastic trajectories for multiple road agents simultaneously. Trajectory prediction for a longer duration would be another interesting and challenging task.

\bibliographystyle{IEEEtran}
\bibliography{main}

\end{document}